\documentclass[journal]{IEEEtran}

\usepackage{graphicx}
\usepackage{amsmath,amssymb} % define this before the line numbering.
\usepackage{cite}
\usepackage{hyperref}
\newcommand{\figdir}{figures}
\def\swone{0.98\linewidth}

\def\swthree{0.31\linewidth}
\def\swfour{0.24\linewidth}

\def\swseven{0.14\linewidth}
\def\sweight{0.12\linewidth}
\def\vspacesection{\vspace{-3mm}}
\def\M{\textbf{M}}
\def\N{\textbf{N}}
\def\H{\textbf{H}}
\newcommand{\reffig}[1]{Fig.\ref{#1}}
\newcommand{\refeq}[1]{Eq.\ref{#1}}
\newcommand{\reftable}[1]{Table.\ref{#1}}

\begin{document}

\title{3D Hand Pose Tracking and Estimation Using Stereo Matching}

\author{Jiawei~Zhang,
        Jianbo~Jiao,
        Mingliang~Chen,
        Liangqiong~Qu,
        Xiaobin~Xu,
        and~Qingxiong~Yang
\thanks{J. Zhang (\href{mailto:jiawzhang8-c@my.cityu.edu.hk}{jiawzhang8-c@my.cityu.edu.hk}), J. Jiao (\href{mailto:Jambol.Jiao@my.cityu.edu.hk}{Jambol.Jiao@my.cityu.edu.hk}), M. Chen (\href{mailto:mlchen2@my.cityu.edu.hk}{mlchen2-c@my.cityu.edu.hk}),  L. Qu, X. Xu (\href{mailto:xiaobinxu2-c@my.cityu.edu.hk}{xiaobinxu2-c@my.cityu.edu.hk}) and Q. Yang (\href{mailto:qiyang@cityu.edu.hk}{qiyang@cityu.edu.hk}) are with Department of Computer Science, City University of Hong Kong, Hong Kong.}
\thanks{L. Qu (\href{mailto:quliangqiong@sia.cn}{quliangqiong@sia.cn}) is also with the State Key Laboratory of Robotics, Shenyang Institute of Automation, Chinese Academy of Sciences, Shenyang,
110016, and the University of Chinese Academy of Sciences, Beijing.}
}% <-this % stops a space

% The paper headers
\markboth{Journal of \LaTeX\ Class Files,~Vol.~14, No.~8, October~2016}%
{Shell \MakeLowercase{\textit{et al.}}: Bare Demo of IEEEtran.cls for IEEE Journals}

% make the title area
\maketitle

% As a general rule, do not put math, special symbols or citations
% in the abstract or keywords.
\begin{abstract}
3D hand pose tracking/estimation will be very important in the next generation of human-computer interaction. Most of the currently available algorithms rely on low-cost active depth sensors. However, these sensors can be easily interfered by other active sources and require relatively high power consumption. As a result, they are currently not suitable for outdoor environments and mobile devices. This paper aims at tracking/estimating hand poses using passive stereo which avoids these limitations. A benchmark\footnote{Stereo hand pose data set can be downloaded from \href{https://sites.google.com/site/zhjw1988/}{https://sites.google.com/site/zhjw1988/}} with 18,000 stereo image pairs and 18,000 depth images captured from different scenarios and the ground-truth 3D positions of palm and finger joints (obtained from the manual label) is thus proposed. This paper demonstrates that the performance of the state-of-the-art tracking/estimation algorithms can be maintained with most stereo matching algorithms on the proposed benchmark, as long as the hand segmentation is correct. As a result, a novel stereo-based hand segmentation algorithm specially designed for hand tracking/estimation is proposed. The quantitative evaluation demonstrates that the proposed algorithm is suitable for the state-of-the-art hand pose tracking/estimation algorithms and the tracking quality is comparable to the use of active depth sensors under different challenging scenarios.
\end{abstract}

% Note that keywords are not normally used for peerreview papers.
\begin{IEEEkeywords}
hand pose tracking/estimation.
\end{IEEEkeywords}

\vspacesection
\section{Introduction}
Vision-based hand pose tracking and estimation will be essential for the next generation of human-computer interaction (HCI) systems. It is a very challenging problem and has many difficulties including high-dimensional articulated structure, severe self-occlusion and chromatically uniform appearance. Many approaches have been proposed recently \cite{supancic2015depth, oberweger2015training, tang2015opening, li20153d, sharp2015accurate, oikonomidis2011markerless, oikonomidis2011efficient, oikonomidis2014evolutionary, qian2014realtime, sridhar2013interactive, sridhar2014real, sridhar2015fast, tang2013real, xu2013efficient, tang2014latent, kirac2014hierarchically, liang2014parsing, sun2015cascaded, stenger2006model, tompson2014real}. Most of them rely on low-cost active depth sensors to solve the self-occlusion and cluttered background problems. Nevertheless, these sensors have limitations. They can be interfered by other active sources such as the sun and another depth sensor. An active sensor takes measurements by emitting its own source of energy and measuring the environment's response. The power consumption is thus relatively high and not suitable for mobile devices. On the other hand, passive stereo does not have these limitations. However, its computational cost could be high and the depth estimates are known to be noisy and unstable especially when the scene is lacks of texture. Indeed, most of the indoor scenes contain large textureless regions. These difficulties exclude passive stereo from being adopted in 3D hand pose tracking/estimation.

All available hand pose tracking/estimation data sets \cite{supancic2015depth, qian2014realtime, sridhar2013interactive, tang2013real, sun2015cascaded, tompson2014real} are captured with active depth sensors. To evaluate the performance of passive stereo for hand pose tracking and estimation, a new benchmark is proposed in this paper. The data set is simultaneously captured by a Point Grey Bumblebee2 stereo camera and an Intel Real Sense F200 active depth camera. We manually label positions of finger joints and center of palm in depth images. Our benchmark contains six different backgrounds with respect to different difficulties (including texture/textureless, dynamic/static, near/far, highlight/no-highlight, etc.) for hand segmentation or disparity estimation. It is difficult to track hand poses with self-occlusions or global rotations of the hand, and thus we also capture two sequences for every background with respect to two different levels of tracking difficulties. As a result, the proposed benchmark has a total of 12 different sequences and every sequence contains 1,500 stereo pairs from Point Grey Bumblebee2 and 1,500 depth images from Intel F200 depth camera. Each sequence is much longer than existing data sets in order to evaluate the long-term tracking performance.

This paper integrates passive stereo with two state-of-the-art hand pose tracking algorithms \cite{oikonomidis2011efficient,qian2014realtime} for quantitative evaluation. Hand pose tracking is known to be slow. It is a significant advance when Qian et al.\cite{qian2014realtime} demonstrates real-time tracking performance on CPU. \cite{oikonomidis2011efficient,qian2014realtime} are both model-based tracking algorithms which search for the optimal parameters that minimize the discrepancy between hypothesized hand model and observed depth image. \cite{oikonomidis2011efficient} uses a stochastic and evolutionary algorithm named Particle Swarm Optimization (\textbf{PSO}) \cite{kennedy2010particle} to seek the estimation from the high dimensional and non-differentiable discrepancy function. However, PSO converges slowly. On the other hand, articulated Iterative Closest Point (\textbf{ICP}) \cite{pellegrini2008generalisation} converges fast but could easily get stuck in the poor local minimal. \cite{qian2014realtime} introduces \textbf{ICPPSO} which takes advantage of the complementary nature of ICP and PSO.

This paper also tests passive stereo for discriminative hand pose estimation. Motivated by the human pose estimation method \cite{shotton2013real}, random forest based hand poses estimation and hand parts labelling methods are emerging \cite{tang2013real, xu2013efficient, tang2014latent, kirac2014hierarchically, liang2014parsing, tang2015opening, li20153d, sun2015cascaded}. Sun et al.'s Cascaded Hand Pose Regression algorithm (\textbf{CHPR}) \cite{sun2015cascaded} is adopted in this paper. \textbf{CHPR} achieves the state-of-the-art estimation accuracy while running at amazing 300fps on a single core CPU. It uses pose indexed features \cite{dollar2010cascaded} to acquire better geometric invariance and utilizes the articulated structure of hand to hierarchically regress the joint positions of hand to make the algorithm more efficient and robust.

Before hand pose tracking/estimation, the hand region should be segmented. An active depth sensor can provide accurate depth information which simplifies the hand segmentation problem. This is not true in stereo matching and turns out to be the major problem in stereo-based hand tracking/estimation.

Skin color information is very useful in hand segmentation. \cite{li2013pixel} uses local appearance features to segment hands from ego-centric videos. Real-time performance is very important in hand tracking. As a result, the efficient color-based skin detection algorithms \cite{terrillon2000comparative, jones2002statistical, vezhnevets2003survey, kakumanu2007survey} are preferred. The existing methods train classifiers from a large amount of skin images. For example, \cite{jones2002statistical} treats skin color as a Gaussian Mixture Model (GMM) and trains a generic model from the abundance images in World Wide Web. However, the performance is still unsatisfied under unconstrained environments (e.g., different lighting conditions and backgrounds). To adapt to different environments, an online training sequence with waving hand for automatic foreground/background segmentation is captured before hand pose tracking/estimation for every scenario in this paper. The adaptive GMM \cite{stauffer1999adaptive, zivkovic2004improved} is then used to perform foreground/background segmentation. The skin and non-skin histogram models can then be computed and the skin color probability is used to segment hand from background.

Stereo matching has been studied for decades and a lot of algorithms have been developed. It uses a pair of images to identify the real distance of every pixel through estimating the projection difference on two cameras. \cite{scharstein2002taxonomy} separates stereo matching algorithms into two categories: local and global algorithms. Local stereo uses filters for matching cost denoising while global stereo explicitly assumes smoothness in the disparity map and minimizes a global cost function. With the assumption of perfect hand segmentation (obtained from manual segmentation), this paper quantitatively evaluates the performance of the state-of-the-art hand pose tracking algorithms using different stereo matching algorithms. Experiments on different local \cite{deriche1993recursively, yoon2006adaptive, tomasi1998bilateral, he2013guided, yang2015recursive, yang2012non} and global \cite{sun2003stereo, bobick1999large} stereo matching methods with different matching costs \cite{kanade1995development, scharstein1994matching, zabih1994non} are conducted. Our experimental results show that the performances of most of the stereo matching methods are comparable to the active depth sensors when the correct hand segmentation is available.
%To preserve the hand edge and computational efficiency, we choose to use guided filter \cite{he2013guided} for hand pose tracking.

However, the traditional passive stereo is much more unstable and noisy than the active sensors in practice. A new stereo algorithm is then proposed for hand tracking. The skin color probability from the online-trained skin color model is used as the guidance of the guided image filter \cite{he2013guided} for matching cost aggregation to increase the robustness around textureless regions. On the other hand, some background regions may have a similar color to skin and thus have high skin probability. Thus a robust hand segmentation method is proposed by using confidence-guided combination of color based hand detection and depth from stereo matching. The experiment results show that the proposed stereo algorithm significantly improves the hand tracking performance. However, it is specially designed for 3D hand pose tracking and may be not suitable for other 3D applications.

The major technical contribution of this paper resides in the proposed stereo-based hand segmentation method Sec. \ref{sec:stereo}. It is indeed a bit similar to traditional sensor fusion works like \cite{yang2007spatial,zhu2008fusion}: an efficient and effective information fusion method but \textbf{specially designed for 3D hand tracking}. Unlike traditional sensor fusion aiming at maintaining the accuracy from all sensors, the proposed method explores how to extract a good hand segmentation by combining the skin color model with passive stereo. This is quite different from the previous works.

The contributions of this paper are summarized as follows:
\begin{enumerate}
    \item a hand pose benchmark with 18,000 stereo image pairs and 18,000 depth images with the ground-truth 3D positions of palm and finger joints;
        %captured from different scenarios
    \item an evidence that commercial passive stereo cameras' performance is comparable to active depth cameras when used for hand pose tracking and estimation;
    \item an evidence that the state-of-the-art hand tracking and estimation methods \cite{oikonomidis2011efficient,qian2014realtime,sun2015cascaded} are robust to passive stereo; and
%    \item a training-based color image hand segmentation method which can adapt to the environment.
    \item a robust stereo-based hand segmentation method specially designed for 3D hand pose tracking and estimation.
\end{enumerate}

%The arrangement of is paper is as follows: We introduce our passive stereo hand pose tracking benchmark in next section. Then we prove that hand pose tracking can achieve comparable performance by traditional stereo matching relative to active depth sensor as long as we can segment hand mask correctly in Sec.\ref{sec:manual_segmentation}. Since it is difficult to get correct hand mask for passive stereo with traditional stereo matching, we propose our stereo method in Sec.\ref{sec:hand_tracking} and experiments in Sec.\ref{sec:experiment} prove our proposed stereo outperforms to traditional stereo and has similar performance relative to active depth sensor for hand pose tracking.

%\iffalse
%\begin{figure}[t]
%\begin{center}
%    \begin{tabular}{cc}
%        \includegraphics[width=\swtwo]{\figdir/disp_vs_depth/B9DY_left_934} &
%        \includegraphics[width=\swtwo]{\figdir/disp_vs_depth/B9DY_right_934}\\
%        \small{(a) Stereo left image}&\small{(b) Stereo right image}\\
%        \includegraphics[width=\swtwo]{\figdir/disp_vs_depth/B9DY_disp_934}&
%        \includegraphics[width=\swtwo]{\figdir/disp_vs_depth/B9DY_depth_934} \\
%        \small{(c) Disparity map}&\small{(d) Depth image}\\
%        \end{tabular}
%\end{center}
%   \caption{(a) and (b) are stereo left and right image from bumblebee camera, (c) estimated disparity map and (d) depth image from SoftKinetic.}
%\label{fig:disp_vs_depth}
%\end{figure}
%\fi

\vspacesection

\section{Stereo-based hand tracking benchmark}\label{sec:benchmark}

\renewcommand{\tabcolsep}{3 pt}
\begin{table}[h]
\caption{Background characteristics in the proposed benchmark.}
\vspace{-1mm}
\centering
\begin{tabular}{|c|c|c|c|c|c|c|}
\hline
               & \textit{B1}                                                        & \textit{B2}     & \textit{B3}                                                          & \textit{B4}      & \textit{B5}      & \textit{B6}      \\ \hline
static/dynamic & static                                                    & static & static                                                      & dynamic & dynamic & dynamic \\ \hline
far/near       & near                                                      & far    & near                                                        & far     & far     & far     \\ \hline
note           & \begin{tabular}[c]{@{}c@{}}rich in\\ texture\end{tabular} &        & \begin{tabular}[c]{@{}c@{}}contain\\ highlight\end{tabular} &         &         &         \\ \hline
\end{tabular}
%\vspace{-10mm}
\label{label:diff_background}
\end{table}
\vspace{-5mm}

\renewcommand{\tabcolsep}{0.5 pt}
\begin{figure}[!ht]%\vspace{-10mm}
    \begin{center}
    \begin{tabular}{ccc}
        \includegraphics[width=\swthree]{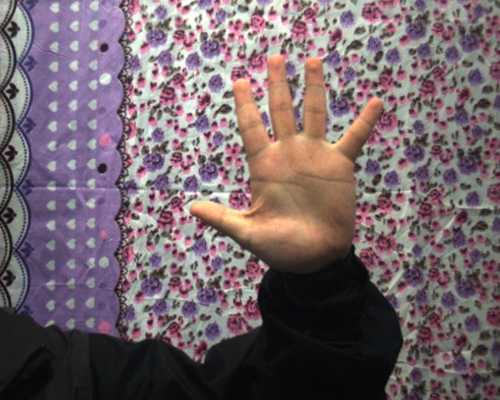}&
        \includegraphics[width=\swthree]{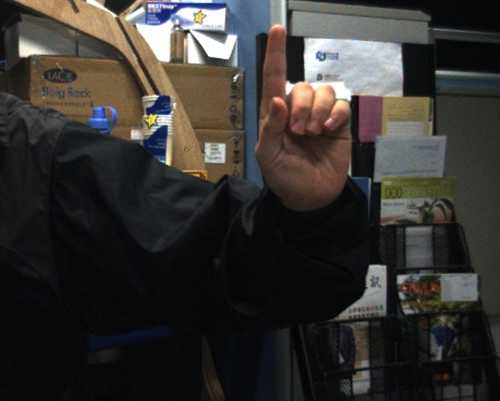}&
        \includegraphics[width=\swthree]{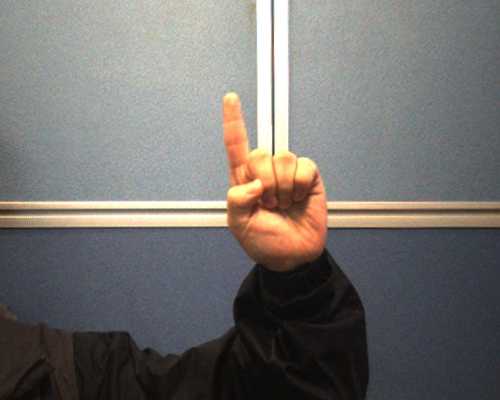}\\
        \small{(a) \textit{B1}}&\small{(b) \textit{B2}}&\small{(c) \textit{B3}}\\
        \includegraphics[width=\swthree]{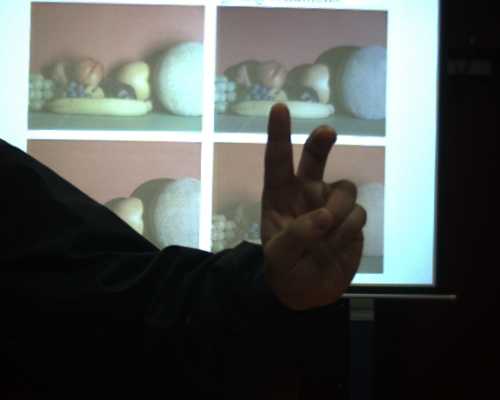}&
        \includegraphics[width=\swthree]{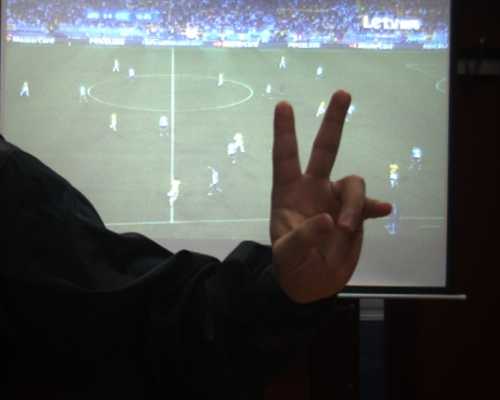}&
        \includegraphics[width=\swthree]{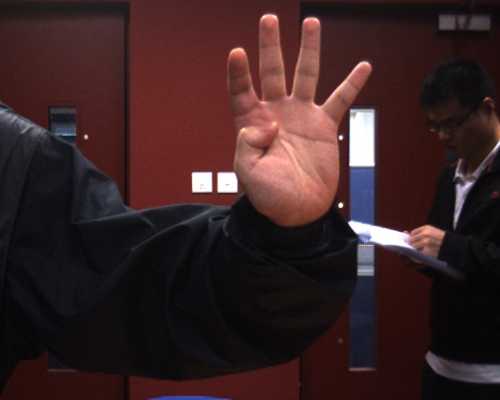}\\
        \small{(d) \textit{B4}}&\small{(e) \textit{B5}}&\small{(f) \textit{B6}}
        \end{tabular}
    \end{center}	
    %\vspace{-5mm}
    \caption{Six different backgrounds are adopted in the proposed benchmark.}
    %\vspace{-5mm}
    \label{fig:diffback}
\end{figure}
\vspace{-5mm}

\renewcommand{\tabcolsep}{0.5 pt}

\begin{figure}[!ht]
    \begin{center}
    \begin{tabular}{ccccc}
        \rotatebox{90}{\small{(a)Counting}}&
        \includegraphics[width=\swfour]{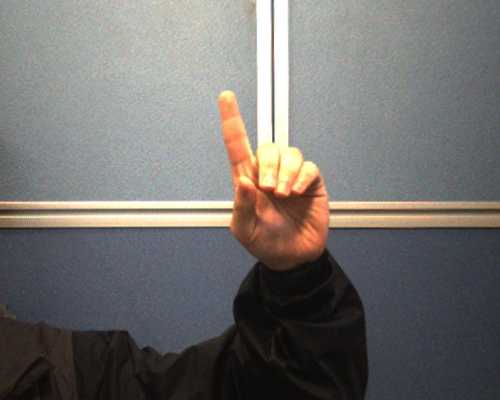}&
        \includegraphics[width=\swfour]{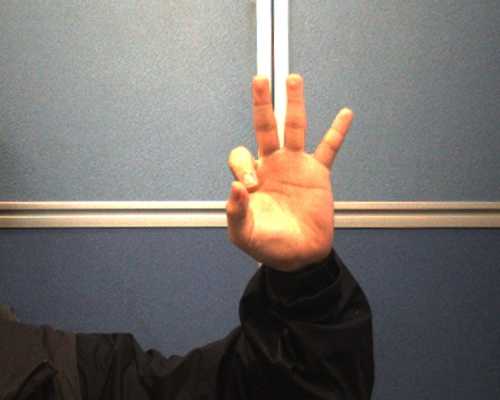}&
        \includegraphics[width=\swfour]{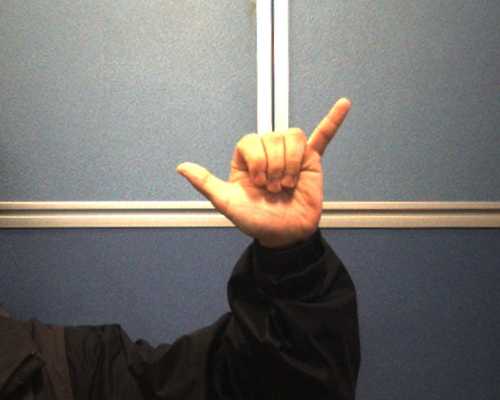}&
        \includegraphics[width=\swfour]{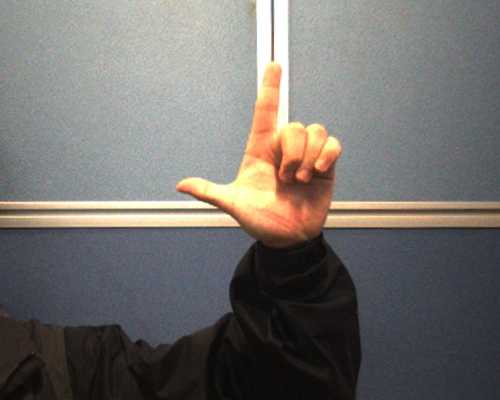}\\
        \rotatebox{90}{\small{(b)Random}}&
        \includegraphics[width=\swfour]{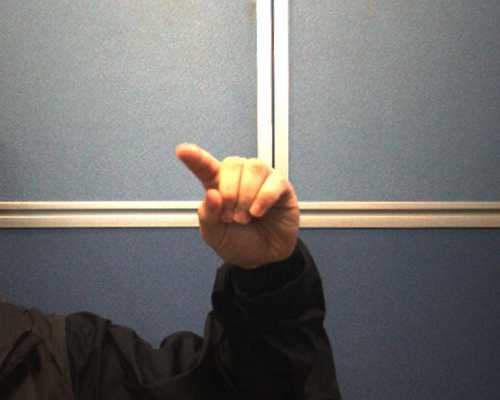}&
        \includegraphics[width=\swfour]{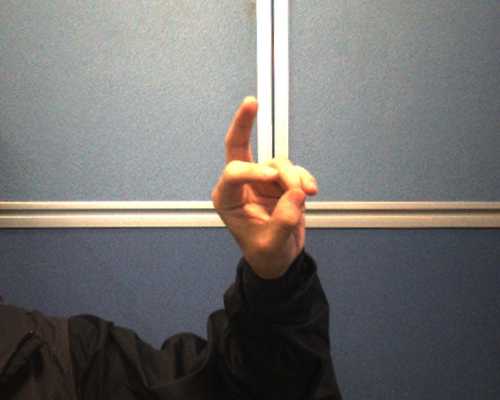}&
        \includegraphics[width=\swfour]{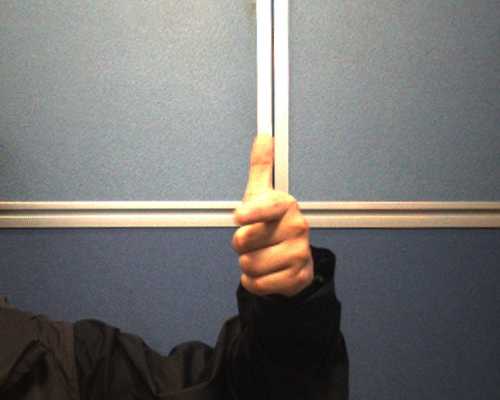}&
        \includegraphics[width=\swfour]{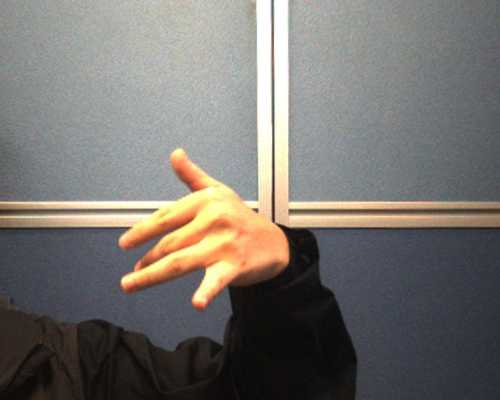}
    \end{tabular}
    \end{center}
    \caption{Two different types of hand poses are adopted in the proposed benchmark: (a) counting/simple and (b) random/difficult poses.}
    \label{fig:diffpose}
\end{figure}

As shown in \reffig{fig:diffback}, the proposed benchmark contains sequences with 6 different backgrounds to evaluate stereo-based hand pose tracking/estimation difficulties in real-world scenarios.

%\renewcommand{\tabcolsep}{2.5 pt}
%\begin{table}[!ht]
%\vspace{-5mm}
%\centering
%\begin{tabular}{|c|c|c|c|c|c|c|}
%\hline
%              & \textit{B1}&\textit{B2}&\textit{B3}&\textit{B4}&\textit{B5}&\textit{B6}\\ \hline
%static/dynamic&static&static&static&dynamic&dynamic&dynamic   \\ \hline
%far/near    &near&far&near&far&far&far   \\ \hline
%note    &rich in texture&&contain highlight&&&   \\ \hline
%\end{tabular}
%\caption{Background characteristics.}
%\vspace{-10mm}
%\label{label:diff_background}
%\end{table}

The characteristics of the adopted backgrounds are summarized in \reftable{label:diff_background}. Most of the indoor backgrounds are textureless which significantly increase the difficulty for passive stereo. Potential highlight and shadow are also quite challenging for both stereo matching and skin color modeling. In addition to static backgrounds, we also capture six sequences with dynamic backgrounds including PowerPoint presentation (\textit{B4}), video playing (\textit{B5}) and people walking (\textit{B6}). If the background is close to the foreground (hand), the depth difference between them is relatively small and thus contributes difficulties for stereo-based hand segmentation.
To adapt to different environments, a training sequence is captured for every background and used for online skin color model training.
%The backgrounds will remain static in all the training sequences except for \textit{B5}. Although dynamic background
%as adaptive GMM based background modeling method do not work well if background is dynamic.

Since self-occlusion and global rotation are two major challenges in 3D hand pose tracking, we capture two sequences for every background with two different poses as shown in \reffig{fig:diffpose}. The first sequence captures simple counting poses with slowly moving fingers as can be seen from \reffig{fig:diffpose} (a). The other sequence is supposed to be much more difficult for hand pose tracking. The hand/fingers move randomly with severe self-occlusions and global rotations as shown in \reffig{fig:diffpose} (b). It is also difficult for a real-time stereo matching algorithm to obtain highly-accurate disparity estimates around slanted fingers and palm in this sequence. Counting and random poses are designed to be similar for all the six different backgrounds to ensure a fair comparison.
%But they actually have a little differences.

For quantitative comparison, the stereo and depth images are captured from a Point Grey Bumblebee2 stereo camera and an Intel Real Sense F200 active depth camera simultaneously. Camera calibration\cite{bouguet2004camera} is performed in advance to obtain the intrinsic and extrinsic parameters of the cameras. The ground-truth positions of finger joints and palm centers are obtained from manual labeling. Each sequence contains 1,500 frames. It is longer than the other 3D hand pose tracking/estimation data sets and our benchmark has a total of 12 sequences.

\vspacesection
\renewcommand{\tabcolsep}{0.5 pt}
\begin{figure*}[!ht]
%\vspace{-5mm}
\begin{center}
    \begin{tabular}{ccc}
        \includegraphics[width=\swthree]{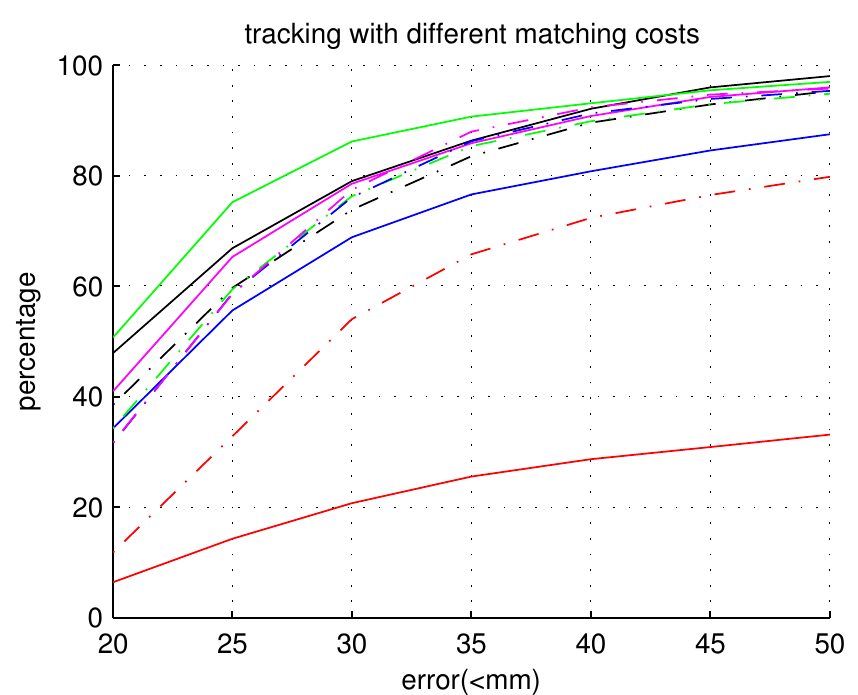} &
        \includegraphics[width=\swthree]{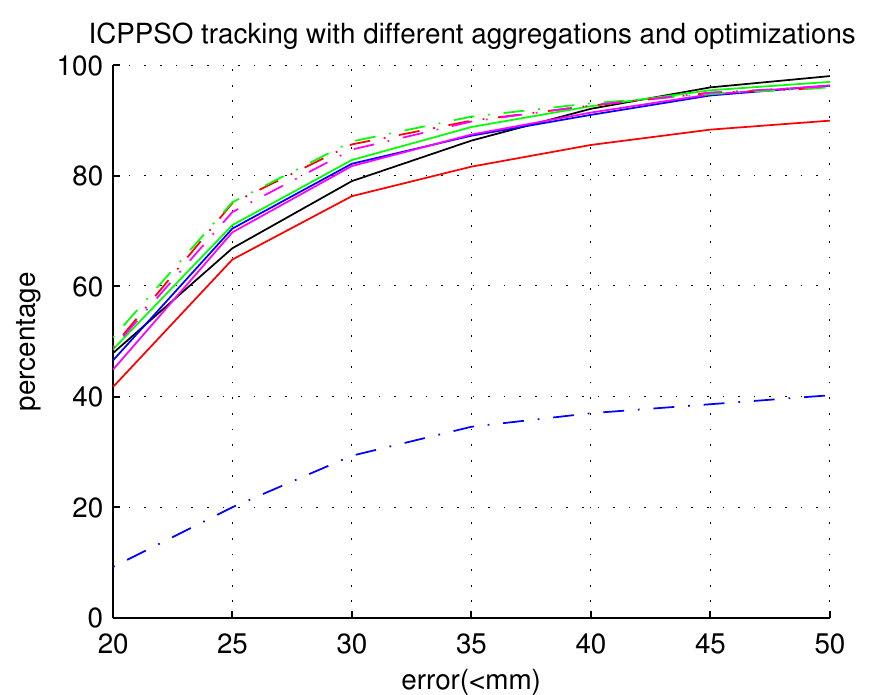} &
        \includegraphics[width=\swthree]{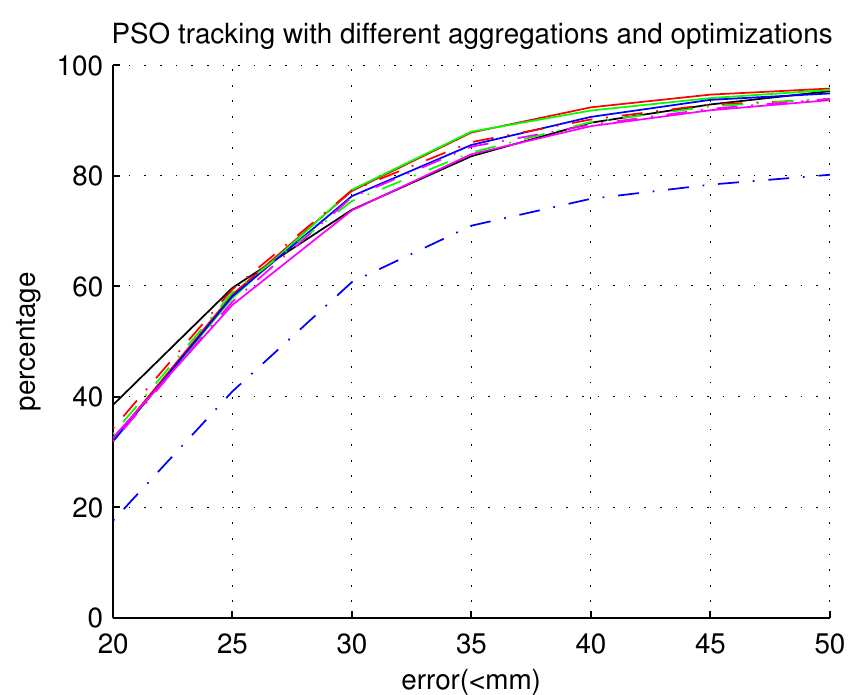}\\
        \includegraphics[width=\swthree]{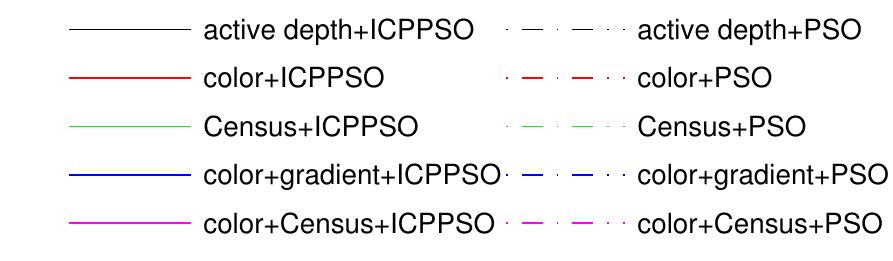} &
        \includegraphics[width=\swthree]{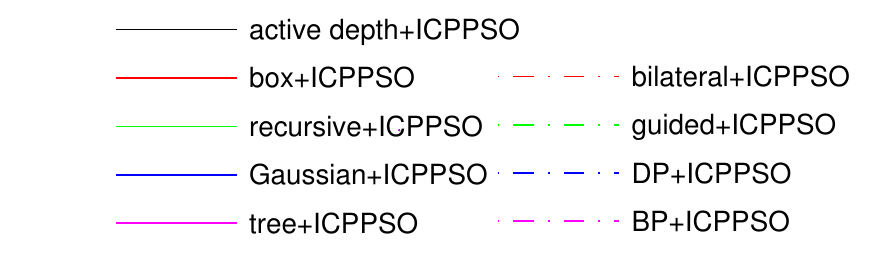} &
        \includegraphics[width=\swthree]{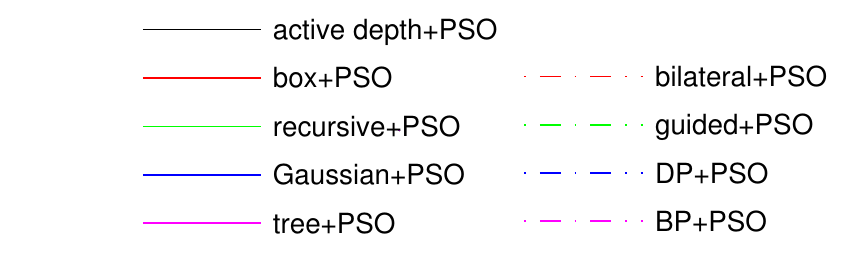}\\
        \small{(a)} & \small{(b) ICPPSO \cite{qian2014realtime}}& \small{(c)PSO \cite{oikonomidis2011efficient}}\\
        {Different matching costs}&\multicolumn{2}{c}{Different cost aggregations \& disparity optimizations}
        \end{tabular}
\end{center}
   %\vspace{-5mm}
   \caption{Percentage of all joints that have a maximum error less than x mm (where x starts from$^{\text{\ref{fn:20mm}}}$ 20) for different stereo matching algorithms with manually labeled hand mask. (a) is the percentages with respect to different types of matching cost. The best cost aggregation or disparity optimization methods are adopted for all the sequences in this experiment. \textit{Note that Census transform\cite{zabih1994non} (which corresponds to the green curves) outperforms the others on average}. (b) and (c) are the percentages with respect to different cost aggregation and disparity optimization methods when Census transform is adopted for matching cost computation. (b) is obtained from ICPPSO \cite{qian2014realtime} and (c) is from PSO \cite{oikonomidis2011efficient} algorithms. \textbf{The dark curves in (a)-(c) represent the percentages obtained with an active depth camera (F200)}. As can be seen from (b)-(c), \textit{the performance of active depth camera is close to most cost aggregation or disparity optimization methods}. }
   %\vspace{-5mm}
\label{fig:100to500_result_counting_random}
\end{figure*}

\section{Tracking with manual hand segmentation}\label{sec:manual_segmentation}
This section demonstrates that the performance of the state-of-the-art hand pose tracking algorithms can be maintained with most stereo matching algorithms on the proposed benchmark, as long as the hand segmentation is correct.

\cite{scharstein2002taxonomy} divides stereo matching into four steps: matching cost computation, cost aggregation, disparity optimization and disparity refinement. In this section, four different matching costs including color difference \cite{kanade1995development}, Census transform \cite{zabih1994non}, a linear combination of color difference and image gradient difference, a linear combination of color difference and Census transform are tested. Local stereo is focused on cost aggregation for matching cost denoising. Six local stereo methods with respect to six filters including box filter, recursive filter \cite{deriche1993recursively}, Gaussian filter, tree filter \cite{yang2012non}, bilateral filter \cite{tomasi1998bilateral} and guided image filter \cite{he2013guided} are tested. Global stereo is focused on disparity optimization. Two global stereo methods including Belief Propagation (BP) \cite{sun2003stereo} and Dynamic Programming (DP) \cite{bobick1999large} are also tested. By combining different matching costs with cost aggregation or disparity optimization, we evaluate a total of 32 different stereo methods. The parameters of each method are trained on the Middlebury Stereo data set    \cite{scharstein2006middlebury}.
%\textbf{The state-of-art stereo methods are not considered in this paper since they are far away from been real-time and not suitable for hand pose tracking. In addition, \cite{scharstein2006middlebury} shows that there is no significant difference between state-of-art methods and the recent real-time methods like the guided filter \cite{he2013guided}}.

Both PSO \cite{oikonomidis2011efficient} and ICPPSO \cite{qian2014realtime} hand pose tracking algorithms are stochastic algorithms, and thus every sequence is tested five times to reduce randomness. An initial hand segment (in the color image) is obtained automatically from active depth sensor followed by manual refinement. The manual refinement is completed by a group of students and it takes over one week. As a result, correct hand segmentation is guaranteed in this experiment.

Same as the other 3D hand pose tracking methods, the percentages of all estimated joints (including palm center, 15 finger joints and 5 finger tips) that have a max-error less than a threshold are used to evaluate the tracking performance. Error thresholds from 20mm to 50mm\footnote{\textbf{According to \cite{supancic2015depth}, max-error of 20mm is close to human accuracy for nearby hands and 50mm is a roughly correct pose}.\label{fn:20mm}} are adopted in all the conducted experiments in this paper and the percentages of all estimated joints are presented in \reffig{fig:100to500_result_counting_random}.

As can be seen from \reffig{fig:100to500_result_counting_random}(a), the use of color difference as the matching cost has the lowest performance mainly because it is not robust to the inconsistent responses from two different sensors. Meanwhile, Census transform \cite{zabih1994non} is invariant to illumination changes and outperforms the other types of matching cost on average on the proposed real-world benchmark. \reffig{fig:100to500_result_counting_random}(b)-(c) show that most cost aggregation and disparity optimization methods have similar performance. Note that manually-segmented hand masks are used in all these experiments. \reffig{fig:100to500_result_counting_random}(b)-(c) show that dark curves are very close to the others which indeed demonstrates that the hand region reconstruction accuracy of most stereo algorithms is comparable to the adopted active depth camera, as long as the hand is segmented correctly.
%Manually-segmented hand masks will not be available in practice. In this case, edge-preserving filters (e.g., tree filter, bilateral filter and guided image filter) and BP should have a higher performance.
%\textbf{The guided image filter\cite{he2013guided} is very efficient and suitable for parallel implementation. As a result, it was combined with Census transform for the stereo algorithm proposed in Sec. \ref{sec:hand_tracking}.}

\reffig{fig:100to500_result_counting_random}(b)-(c) also show that ICPPSO has a higher performance than PSO. This is because PSO converges slower than ICPPSO. However, ICP is not robust to depth noises and thus ICPPSO has a lower performance when DP is adopted as the disparity optimization method (blue dashed curves).

\textbf{This section concludes that when used for 3D hand tracking, the performance of passive stereo is comparable to that of active depth sensor (dark curves in \reffig{fig:100to500_result_counting_random}), as long as the hand can be segmented correctly.}

\vspacesection

\section{Stereo-based hand pose tracking and estimation}\label{sec:hand_tracking}
\renewcommand{\tabcolsep}{0.1 pt}
\begin{figure*} [!ht]
    \begin{center}
        \includegraphics[width=\swone]{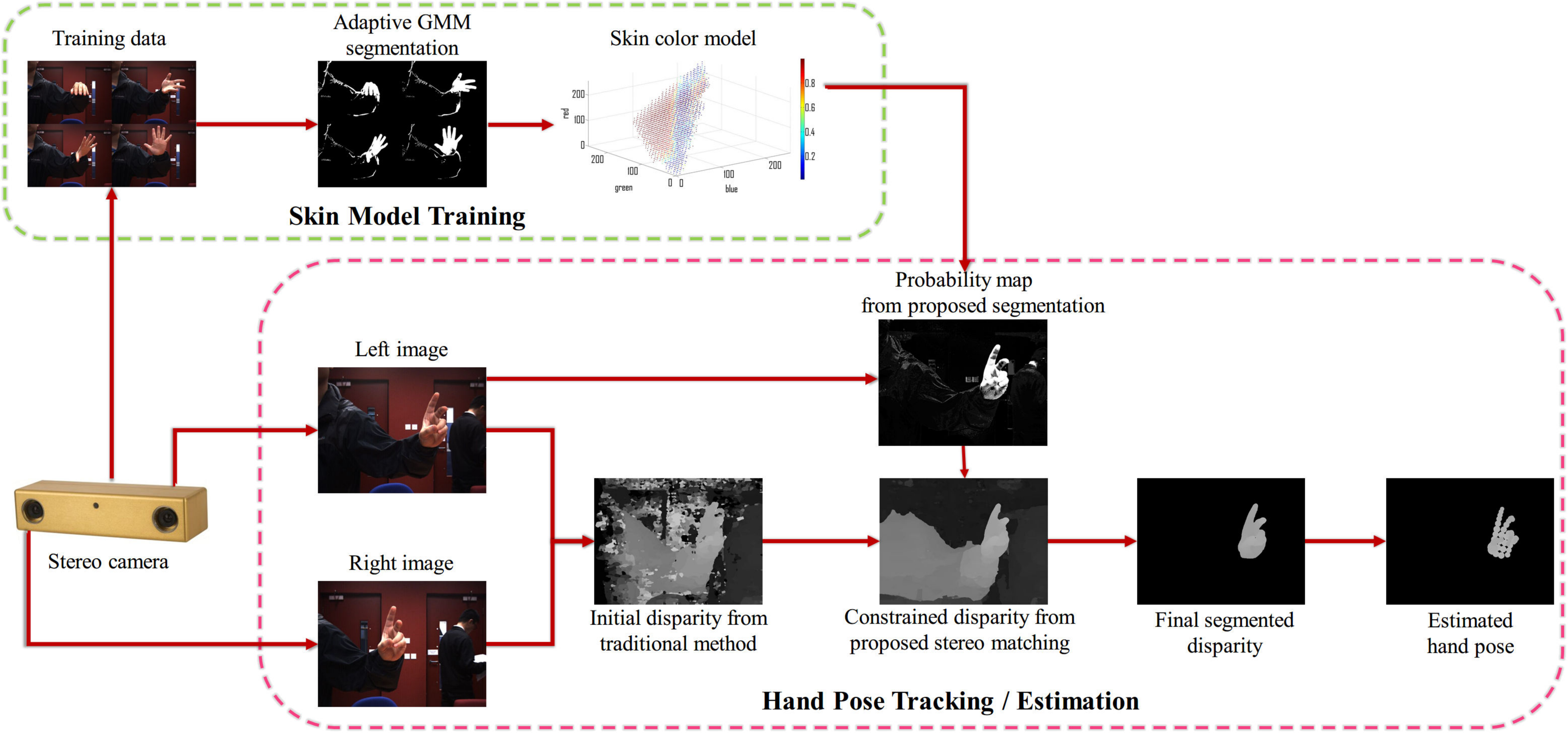}\\
    \end{center}
    \caption{Framework of the proposed stereo based 3D hand pose tracking/estimation method.}
    \label{fig:framework}
\end{figure*}

This section presents the details on the proposed stereo based hand pose tracking/estimation method. The framework is summarized in \reffig{fig:framework}. To adapt to different environments, a training sequence is captured for online training of an adaptive skin color model before hand pose tracking/estimation. During hand pose tracking/estimation, the depth is estimated from a passive stereo. However, traditional stereo matching algorithms are unstable and thus the depth estimates are noisy. A novel constrained stereo matching algorithm specially designed for hand pose tracking/estimation is thus proposed in Sec. \ref{sec:stereo}. The hand region can then be segmented according to the depth estimates and trained skin color model.
%Then we compare the performance between two existing hand pose tracking algorithms \cite{oikonomidis2011efficient, qian2014realtime} and one hand pose estimation algorithms \cite{sun2015cascaded} in the next section.
%The details of our method are shown in the following subsections.

\vspacesection
\subsection{Training based hand modeling}\label{sec:hand_modeling}
As discussed above, hand segmentation should be performed before 3D hand pose tracking/estimation. Unlike the hand segmentation methods adopted with active depth cameras, it is hard to obtain accurate depth from passive stereo. Hand segmentation from color is difficult. Some of the background colors could be similar to skin. In addition, skin color may also vary under different lighting conditions. It is impossible to construct a generic color model which is suitable for all scenarios.

To address these problems, an online training based skin color detector is proposed. A training sequence is captured just before hand pose tracking/estimation. Real-time background modeling methods like adaptive GMM \cite{stauffer1999adaptive, zivkovic2004improved} is adopted to segment foreground object (which we assume to be the hand) from the background.
%Since adaptive GMM will treat static objects as background, we should make sure the hand keeps waving in the training sequence. In addition, we also need the hand to face to most possible directions to make sure the trained color model can work well in the tracking sequence.
The hand should keep waving (for a few seconds) in the training sequence otherwise it cannot be detected as the foreground. This additional training step is reasonable in practice comparing to the other 3D hand tracking methods. For instance, \cite{oikonomidis2011efficient} also requires users to manually fit an initial pose before tracking.

After foreground segmentation, the foreground objects are assumed to be the hand which have a specific skin color. The color histogram for the hands $\H^{hand}$ and the histogram for all the images $\H^{image}$ in the training video sequence are computed. This paper adopts the likelihood ratio approach presented in \cite{jones2002statistical} and uses a threshold for skin color detection. Specifically, the following skin color probability
\begin{equation} \label{eq:P_s_c_est}
    P^{skin}(c)={\H^{hand}(c)}/{\H^{image}(c)}
    %P^{skin}(c)=\frac{\H^{hand}(c)}{\H^{image}(c)}
\end{equation}
is used to obtain an initial hand segmentation from the color images where $c$ represents a color candidate.

\begin{figure*}[!ht]
\begin{center}
    \begin{tabular}{ccccccccc}
        \rotatebox{90}{\textit{B1}}&
        \includegraphics[width=\sweight]{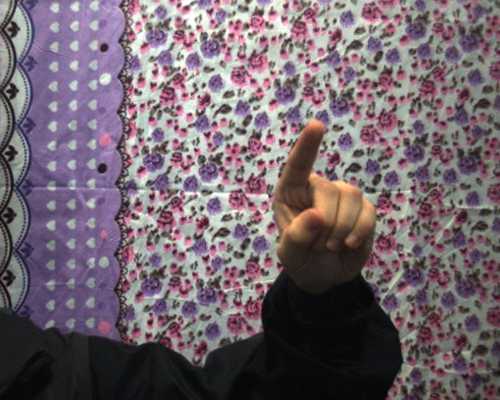} &
        \includegraphics[width=\sweight]{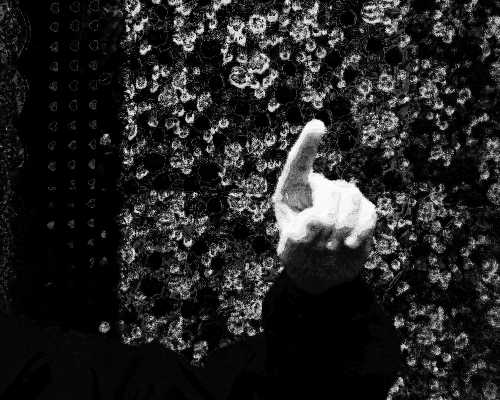} &
        \includegraphics[width=\sweight]{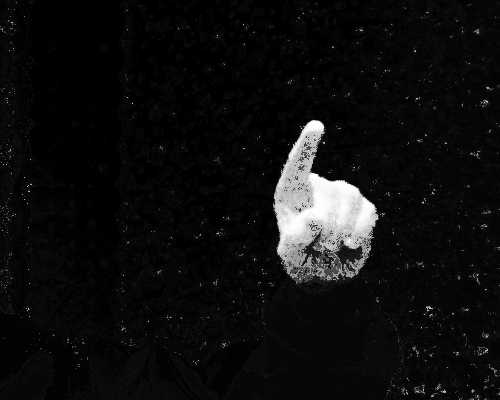} &
        \includegraphics[width=\sweight]{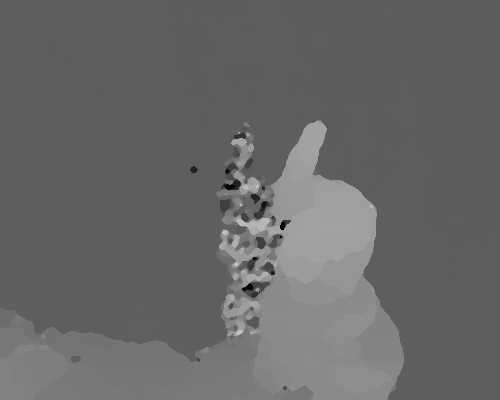} &
        \includegraphics[width=\sweight]{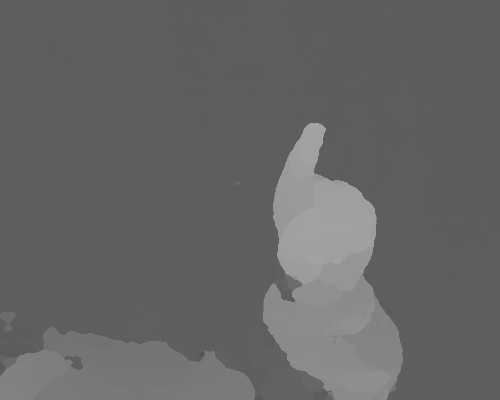} &
        \includegraphics[width=\sweight]{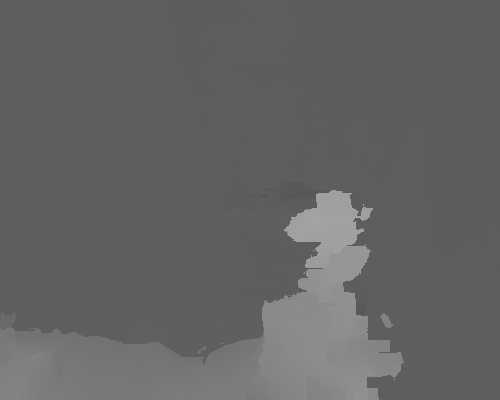} &
        \includegraphics[width=\sweight]{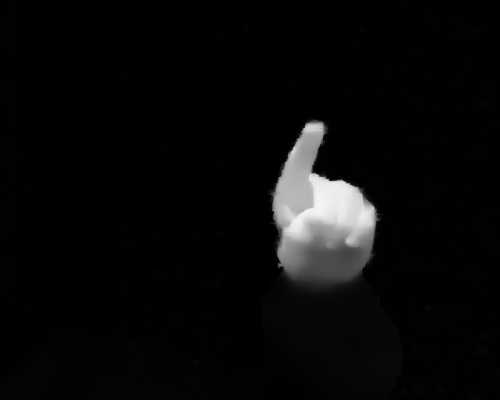} &
        \includegraphics[width=\sweight]{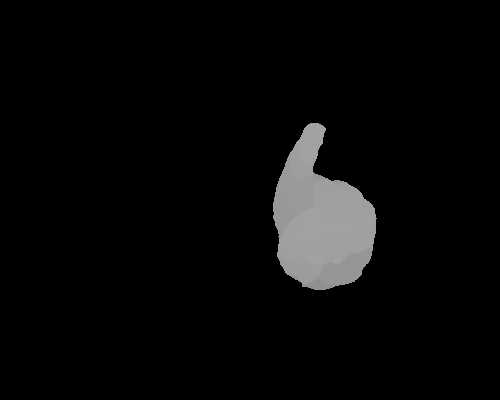}\\
        \rotatebox{90}{\textit{B2}}&
        \includegraphics[width=\sweight]{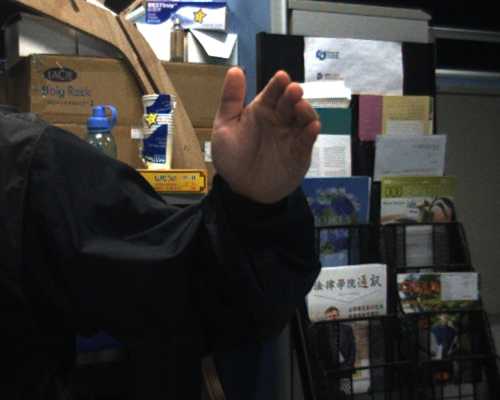} &
        \includegraphics[width=\sweight]{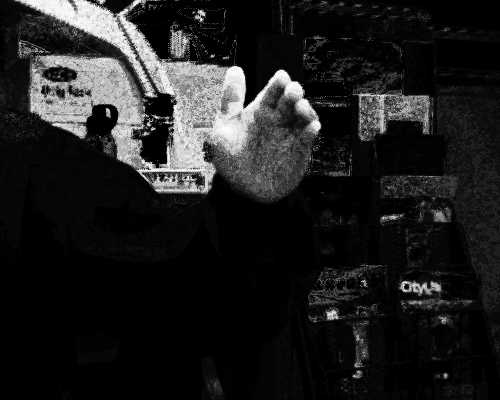} &
        \includegraphics[width=\sweight]{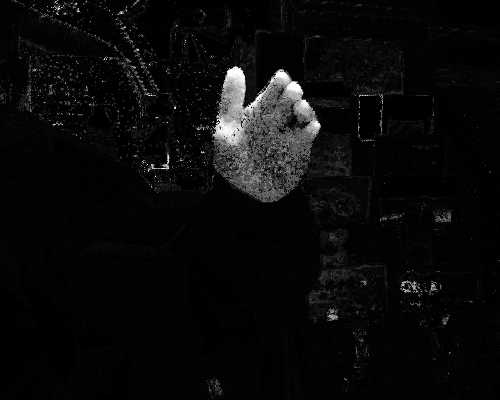} &
        \includegraphics[width=\sweight]{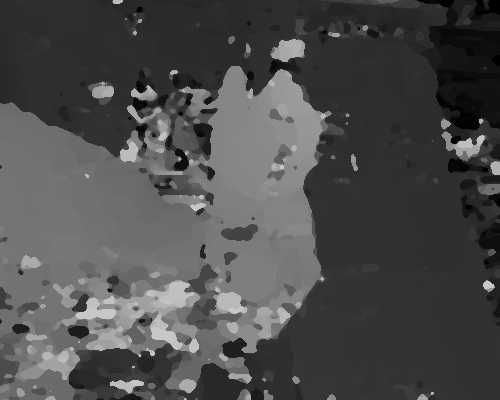} &
        \includegraphics[width=\sweight]{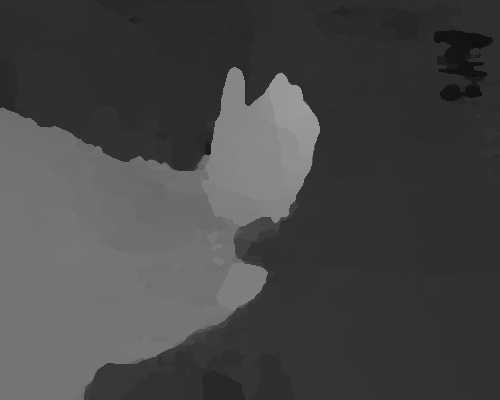} &
        \includegraphics[width=\sweight]{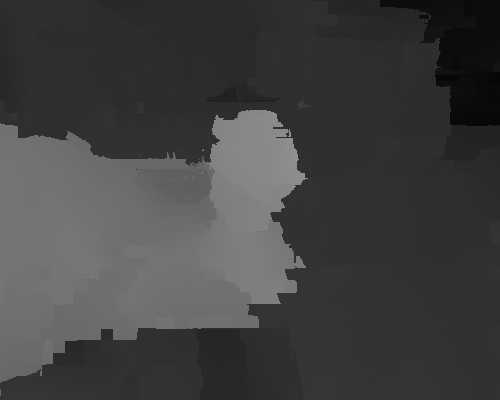} &
        \includegraphics[width=\sweight]{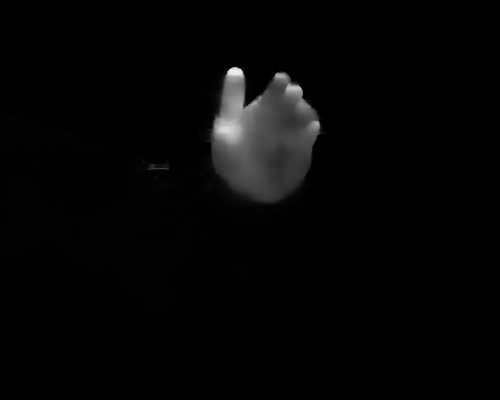} &
        \includegraphics[width=\sweight]{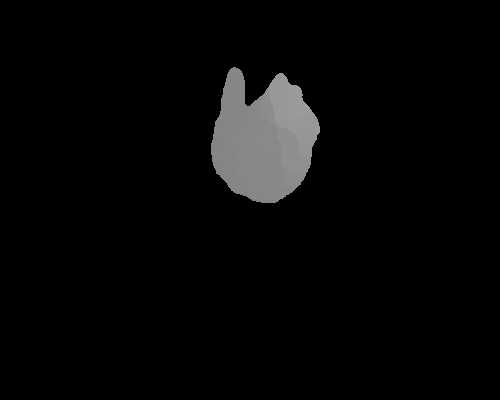}\\
        \rotatebox{90}{\textit{B3}}&
        \includegraphics[width=\sweight]{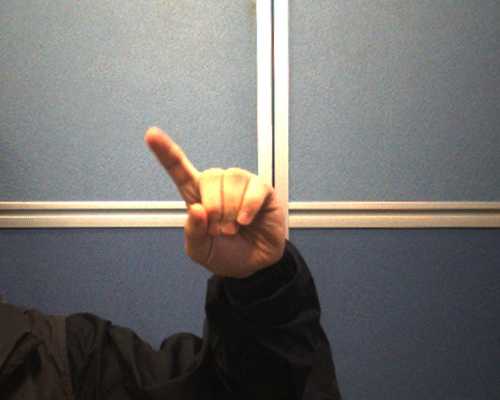} &
        \includegraphics[width=\sweight]{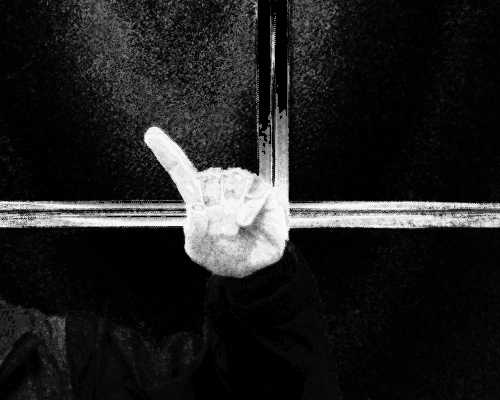} &
        \includegraphics[width=\sweight]{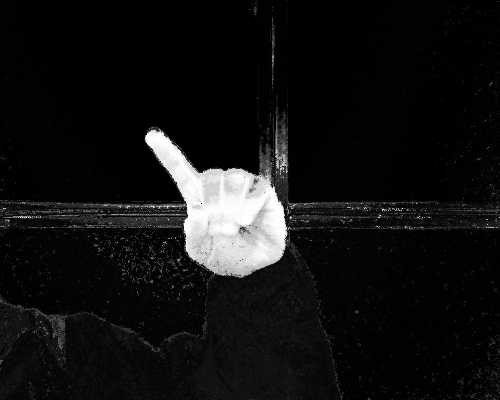} &
        \includegraphics[width=\sweight]{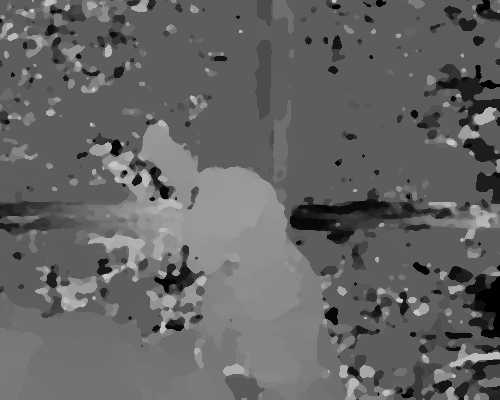} &
        \includegraphics[width=\sweight]{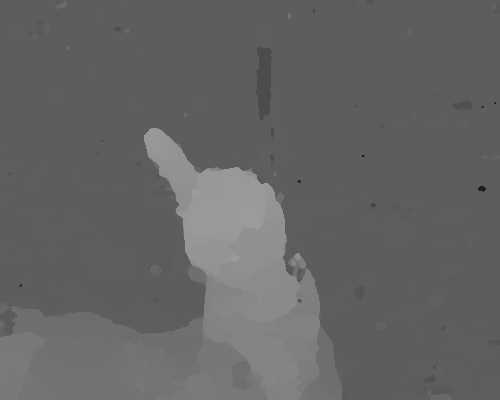} &
        \includegraphics[width=\sweight]{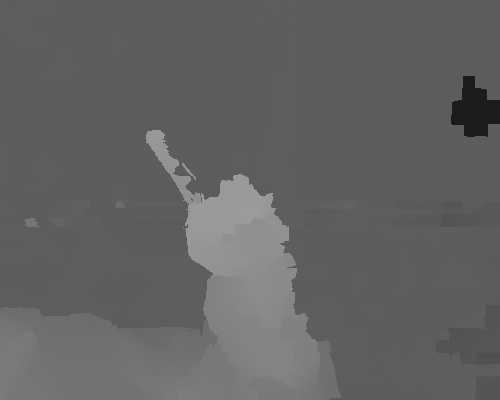} &
        \includegraphics[width=\sweight]{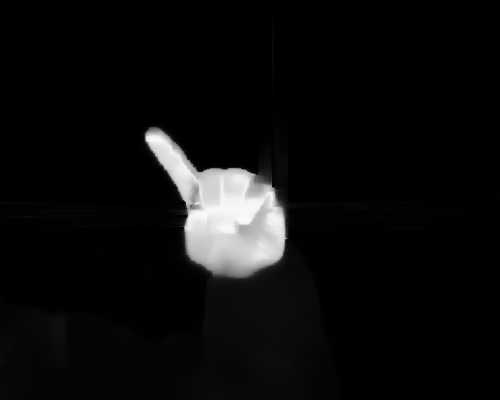} &
        \includegraphics[width=\sweight]{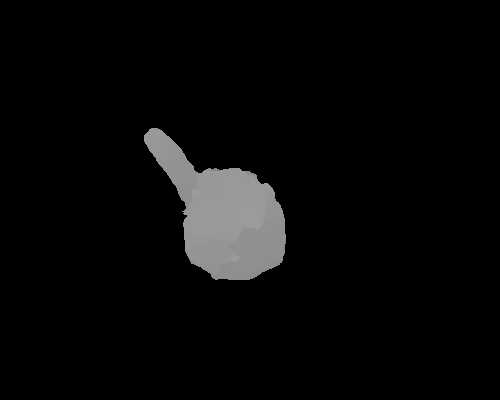}\\
        \rotatebox{90}{\textit{B4}}&
        \includegraphics[width=\sweight]{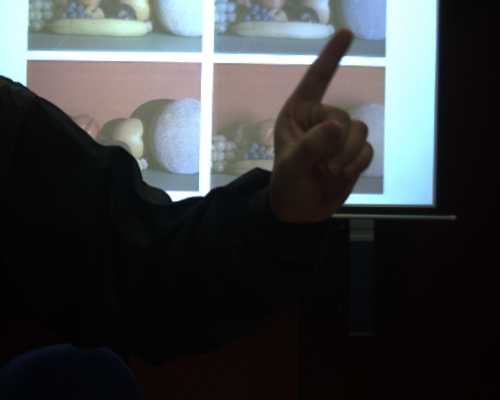} &
        \includegraphics[width=\sweight]{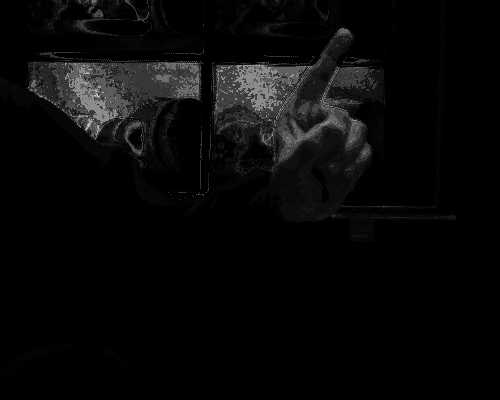} &
        \includegraphics[width=\sweight]{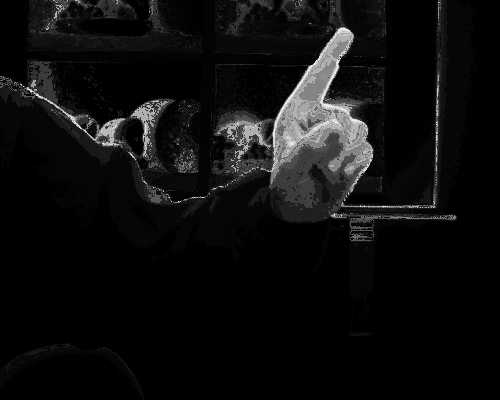} &
        \includegraphics[width=\sweight]{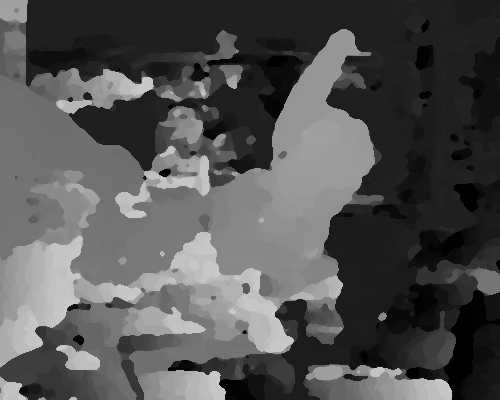} &
        \includegraphics[width=\sweight]{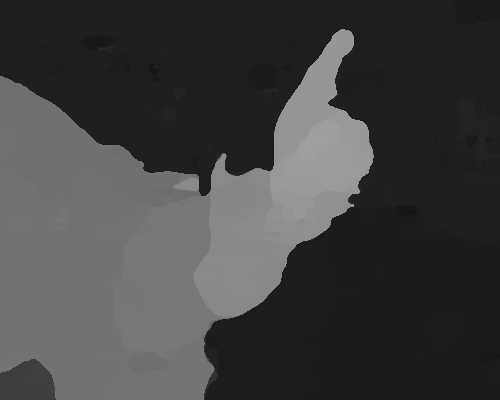} &
        \includegraphics[width=\sweight]{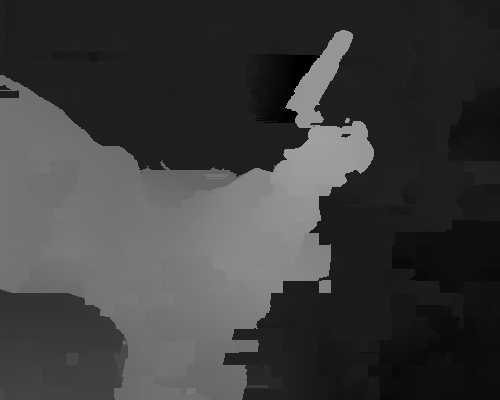} &
        \includegraphics[width=\sweight]{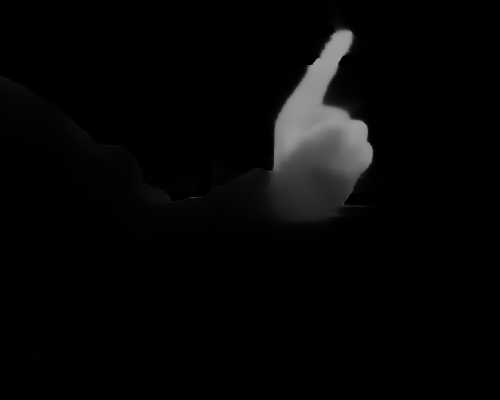} &
        \includegraphics[width=\sweight]{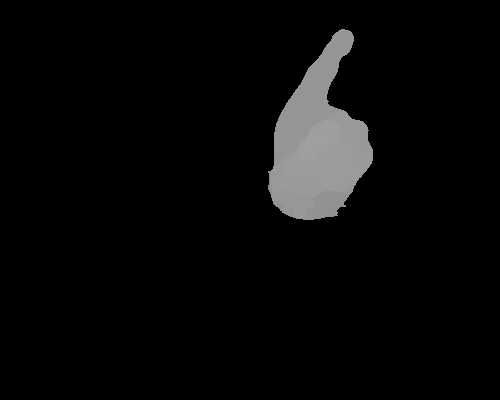}\\
        \rotatebox{90}{\textit{B5}}&
        \includegraphics[width=\sweight]{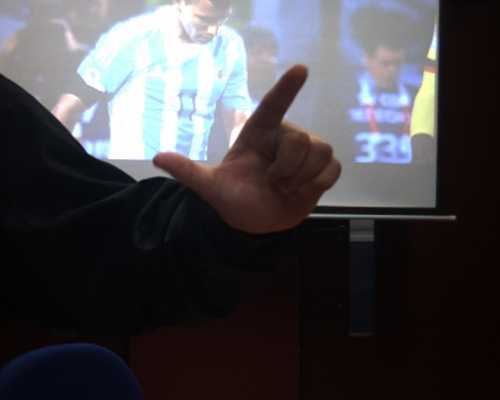} &
        \includegraphics[width=\sweight]{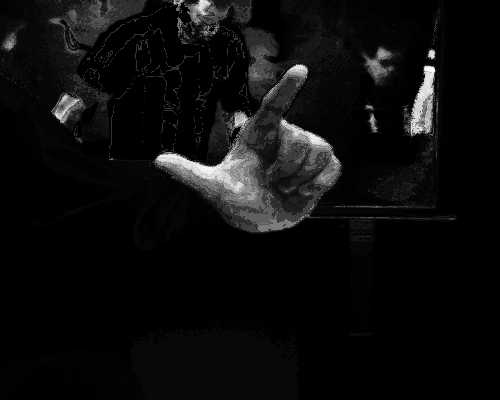} &
        \includegraphics[width=\sweight]{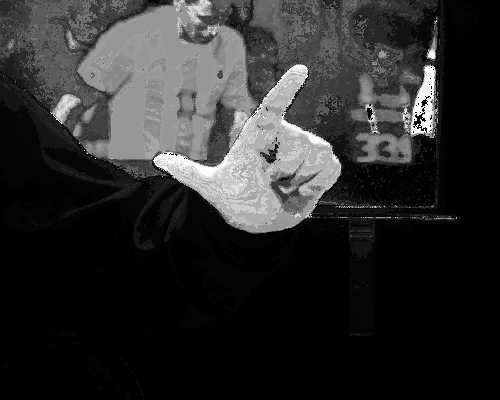} &
        \includegraphics[width=\sweight]{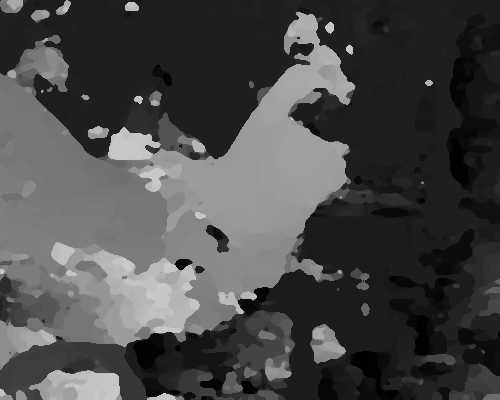} &
        \includegraphics[width=\sweight]{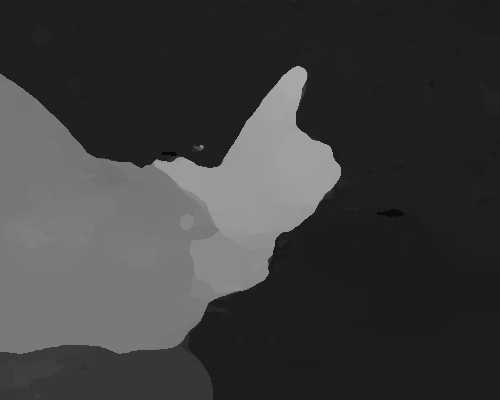} &
        \includegraphics[width=\sweight]{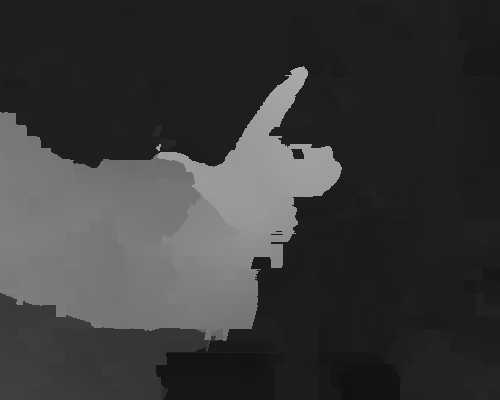} &
        \includegraphics[width=\sweight]{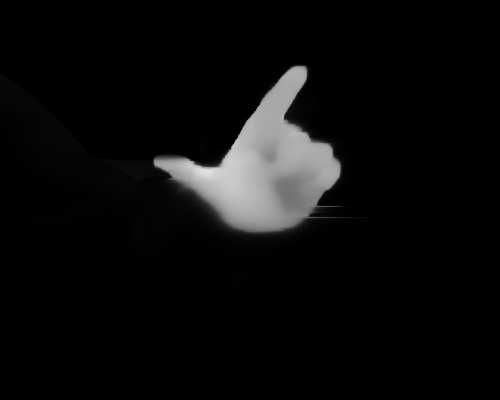} &
        \includegraphics[width=\sweight]{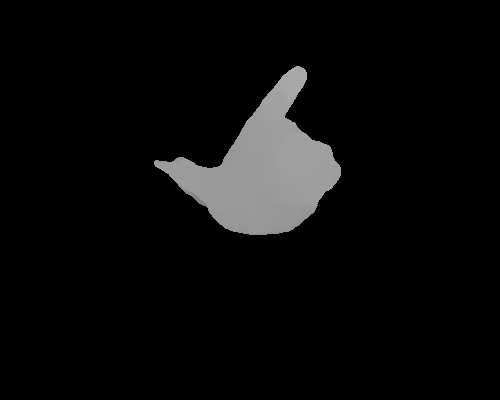}\\
        \rotatebox{90}{\textit{B6}}&
        \includegraphics[width=\sweight]{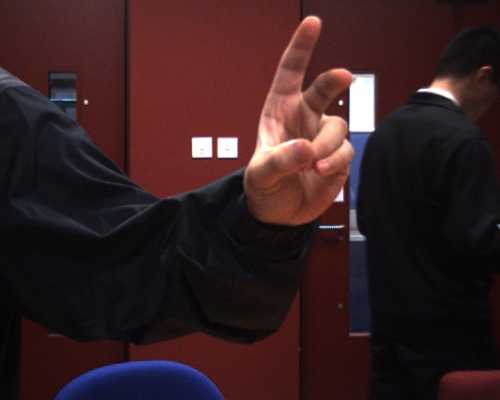} &
        \includegraphics[width=\sweight]{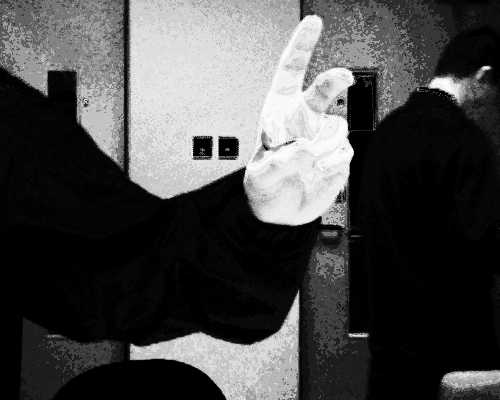} &
        \includegraphics[width=\sweight]{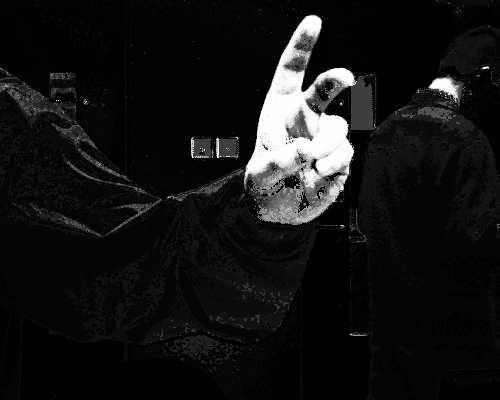} &
        \includegraphics[width=\sweight]{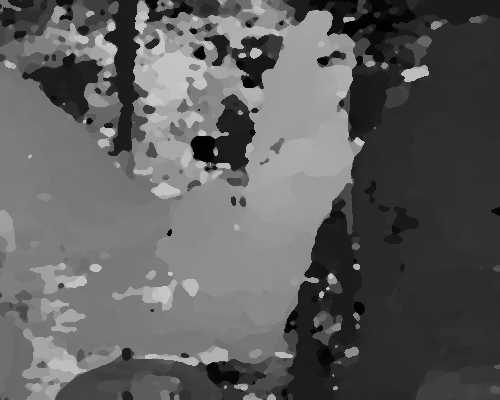} &
        \includegraphics[width=\sweight]{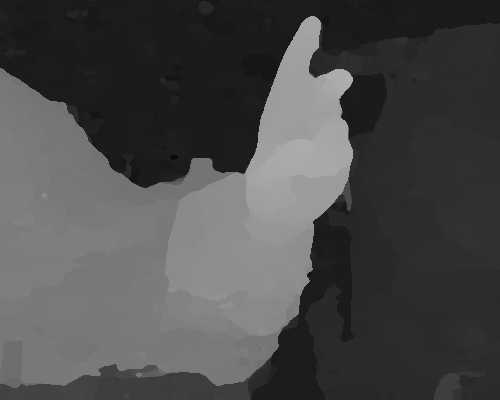} &
        \includegraphics[width=\sweight]{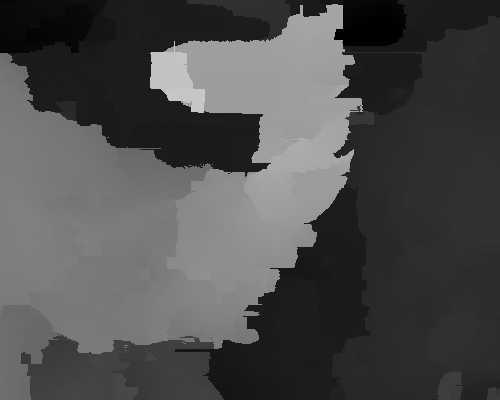} &
        \includegraphics[width=\sweight]{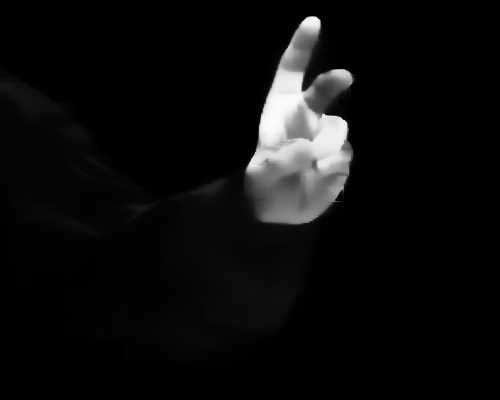} &
        \includegraphics[width=\sweight]{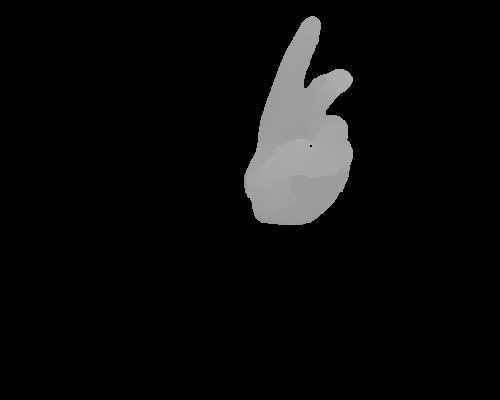}\\
        &\tiny{(a)color}&\tiny{(b)\cite{jones2002statistical}}&\tiny{(c)proposed}&\tiny{(d)traditional}&\tiny{(e)proposed}&\tiny{(f)Meshstereo}&\tiny{(g)Eq.\eqref{eq:P_hand}:}&\tiny{(h)segmented}\\
        &                &\tiny{$P^{skin}(c)$}                      &\tiny{$P^{skin}(c)$}   &\tiny{stereo}        &\tiny{stereo}&\tiny{\cite{zhang2015meshstereo}}     &\tiny{$P^{hand}(c,d)$}             &\tiny{disparity}
%        &(a)&(b)$P(s|c)$&(c)$P(s|c)$&(d)traditional &(e)proposed&(f)$P(g)$          &(h)segmented\\
%        &color&\cite{jones2002statistical}&proposed&stereo    &stereo    &hand probability&hand disparity
        \end{tabular}
\end{center}
   \caption{Intermediate results of the proposed hand tracking/estimation framework. (a) presents the reference color image. (b)-(c) are the skin probability maps obtained from the generic GMM skin color model\cite{jones2002statistical} and the proposed skin color model, respectively. (d) is the disparity maps obtained from a traditional local stereo method with Census transform as the matching cost and guided image filter for cost aggregation. (e) is the disparity maps obtained from the stereo matching algorithm proposed in Sec. \ref{sec:stereo}. (f) is the disparity maps obtained from Meshstereo\cite{zhang2015meshstereo} which is very slow but the top stereo matching method on Middlebury benchmark (without training). The source code (with the default parameters) published by the authors are used. (g) is the hand probability obtained from the hand segmentation method proposed in Sec. \ref{sec:hand_segmentation}. (h) is the segmented hand regions with the estimated disparities.}
\label{fig:mask_comp}
\end{figure*}

\reffig{fig:mask_comp} (b)-(c) compares the skin color probability $P^{skin}(c)$ from \cite{jones2002statistical} and the training method proposed in this section. Different from the proposed method, the generic skin color model in \cite{jones2002statistical} is trained by plenty of images from World Wide Web. \reffig{fig:mask_comp} (b)-(c) show that the proposed model can better separate skin region from the other objects while \cite{jones2002statistical} is more ambiguous especially for red backgrounds. It is simply because the generic skin color model in \cite{jones2002statistical} is trained from a large amount of images, and thus it treats more objects/colors as skin (e.g., \textit{B1}, \textit{B2}, \textit{B3} and \textit{B6} in \reffig{fig:mask_comp} (b)). However, skin colors normally only dominate a small region in the color space for a specific scenario. The generic skin color model \cite{jones2002statistical} provides unsatisfied skin detection result in dark environments like \textit{B4}. It is likely because this type of lighting condition never or seldom appears in its training data set. As a result, generic skin color model assigns very low skin probability to shadows (on the hand). But shadows indeed frequently appear especially for slanted/close hands (e.g., \textit{B1} and \textit{B3} in \reffig{fig:mask_comp} (b)). The proposed hand detector is more robust than \cite{jones2002statistical} mainly because a specific skin color model is trained for every individual scenario. \textit{B5} in \reffig{fig:mask_comp} (b) shows that the hand probability from the proposed method is relatively high in the background. It is because to ensure real-time performance, adaptive GMM background modeling is adopted, and thus the foreground segmentation quality is not very high especially when the background objects have fast motions like \textit{B5}. However, this problem could be solved with the depth information as shown in \reffig{fig:mask_comp} (e)(g)(h). The details will be presented in the following subsections.

\vspacesection
\renewcommand{\tabcolsep}{0.1 pt}

\subsection{Constrained stereo matching}\label{sec:stereo}

A traditional stereo matching algorithm does not make any specific assumption on the scene to be captured. It performs well on textured/synthetic scenes as shown in the first row in Fig.\ref{fig:mask_comp} (d). However, its performance may drop dramatically in a real-world indoor environment where most of the objects contain large textureless regions as can be seen from the last two rows in Fig.\ref{fig:mask_comp} (d).

However, a traditional stereo matching algorithm is not required in this paper. As has been demonstrated in Sec.\ref{sec:manual_segmentation}, most of the existing stereo algorithms perform well when the hand segmentation is correct. This means that \textbf{the objective stereo algorithm only needs to maintain (i) the depth accuracy of the hand and (ii) a clear depth difference between the hand and the background objects.} Unlike the traditional stereo algorithm, the objective stereo algorithm does not require high depth accuracy on these objects as long as the estimated depth is sufficiently far away from the hand. This is similar to the use of short-range active depth sensors. They cannot measure background objects if they are out of the range but they still perform well because these background objects can be directly ignored.

An integrated stereo matching is proposed in this section. It is specially designed for 3D hand tracking with the assumption that the hand skin is visible. The occluded pixels are detected using the left-right consistency check and the unstable pixels (due to lack of texture, specularity, etc.) are detected based on matching cost confidence \cite{ivc-04-egnal}. Matching cost confidence is used to measure how distinct the best peak is. Let $\M_p$ denotes the original matching cost at pixel $p$, and $\M_p^1$ and $\M_p^2$ denote the matching cost for the best and second best disparity/depth values, respectively. The cost confidence is defined as
$|(\M_p^1-\M_p^2)/\M_p^2|$. If it is under 0.04, pixel $p$ is declared unstable.

Let $d$ denote a depth/disparity candidate, a new matching cost $\N_p$ at pixel $p$ that excludes the contribution of occlusions is computed as follows:
\begin{equation}
\N_p(d)=\left\{ \begin{array}{cl}
0 & \textrm{if $p$ is occluded,}  \\
\M_p(d) & \textrm{otherwise.}
\end{array} \right.
\label{eq:new_matching_cost}
\end{equation}
Instead of the reference color image, the skin probability $P^{skin}(c)$ estimated from the model proposed in Sec. \ref{sec:hand_modeling} is used as the guidance image for cost aggregation on the new matching cost and let $\N_p^F$ denote the aggregated cost at $p$. As shown in Fig.\ref{fig:mask_comp}(c), most of the non-skin regions are very dark and thus the guidance image filter kernel is very large around those regions. As a result, the corresponding aggregated cost values are quite stable inside these regions. Let $D_p^\N$ denote the depth obtained from $\N_p^F$ (using the traditional winner-take-all strategy). $D_p^\N$ is normally over-smoothed around non-skin regions due to the adoption of the large filter kernel. However, as analyzed at the beginning of this subsection, the depth accuracy requirement is low on non-skin regions.

Additionally, $D_p^\N$ is the intermediate depth estimation but not the final result. It is only used to adjust the original matching cost:
\begin{equation}
\M_p(d)\leftarrow\left\{ \begin{array}{cl}
\alpha|d-D_p^\N| & \textrm{if $p$ is occluded,}  \\
\M_p(d)+\beta|d-D_p^\N| & \textrm{if $p$ is unstable,}  \\
\M_p(d) & \textrm{otherwise.}
\end{array} \right.
\label{eq:new_matching_cost}
\end{equation}
$\M_p$ is the original matching cost, $\alpha=2$ and $\beta=0.5$ are two constants determining the contribution of the intermediate depth $D_p^\N$ for occluded and other unstable pixels respectively.

The new matching cost is filtered using guided image filter \cite{he2013guided} with the reference color as the guidance to compute the final depth/disparity map in Fig.\ref{fig:mask_comp}(e). Note that most of the noisy depth estimated in Fig.\ref{fig:mask_comp}(d) are removed from Fig.\ref{fig:mask_comp}(e) although the background depth is slightly over-smooth. However, the depth accuracy on the hand region is well-preserved because the filter kernel is relatively small when cost aggregation is performed on $\N_p$ and the pixels inside the hand region are mostly stable pixels. But it should be noted that the stereo matching proposed in this section is specially designed for 3D hand pose tracking and may be not suitable for other 3D applications.

Although the state-of-the-art stereo methods like Meshstereo may also produce ``clean'' backgrounds as can be seen from \reffig{fig:mask_comp} (f), it is very slow and the performance around the hand region is obviously lower than the proposed method.
%our proposed stereo \reffig{fig:mask_comp} (e) can obviously provide better disparity map and it is impossible to use Meshstereo for hand pose tracking. In addition, it will take Meshstereo more than 1 minute to estimate disparity for a image pair which is extremely slow.

\vspacesection

\subsection{Hand segmentation}\label{sec:hand_segmentation}

In this paper, a pixel is inside the hand region if its skin color probability $P^{skin}(c)$ (obtained from \refeq{eq:P_s_c_est}) is high and its depth $d$ is close to the hand depth in the previous frame. Under this assumption, a hand probability for each pixel can be defined as
\begin{equation}\label{eq:P_hand}
P^{hand}(c,d)=P^{skin}(c)N(d;\mu_d,\sigma_d),
\end{equation}
where $N(d;\mu_d,\sigma_d)$ is a Gaussian distribution. The mean $\mu_d$ is the average hand depth in the previous frame and the standard deviation $\sigma_d$ is fixed to 150mm in all the conducted experiments. Finally, a pixel is assumed to be inside the hand region if $P^{hand}(c,d)$ $>$ $0.1$.

Some hand segmentation results with disparities are presented in \reffig{fig:mask_comp}(h). When the background is highly-textured (e.g., \textit{B1}), the disparity from traditional stereo is sufficiently accurate. However, applying traditional stereo on \textit{B6} results in many disparity noises in the background due to the lack of texture. The constrained stereo matching algorithm proposed in Sec. \ref{sec:stereo} can obtain globally-smooth (although may not be very accurate) disparity estimates in the background region and thus is very useful in hand segmentation. For some specific backgrounds like \textit{B5}, although trained skin color model is not good enough (mainly due to incorrect foreground/background segmentation during training), accurate hand segmentation could still be obtained with the help of the disparity estimates from the proposed stereo matching algorithm.

\vspacesection

\section{Experiments}\label{sec:experiment}

\renewcommand{\tabcolsep}{0.1 pt}
\begin{figure*}[!t]
\begin{center}
    \begin{tabular}{ccc}
        \includegraphics[width=\swthree]{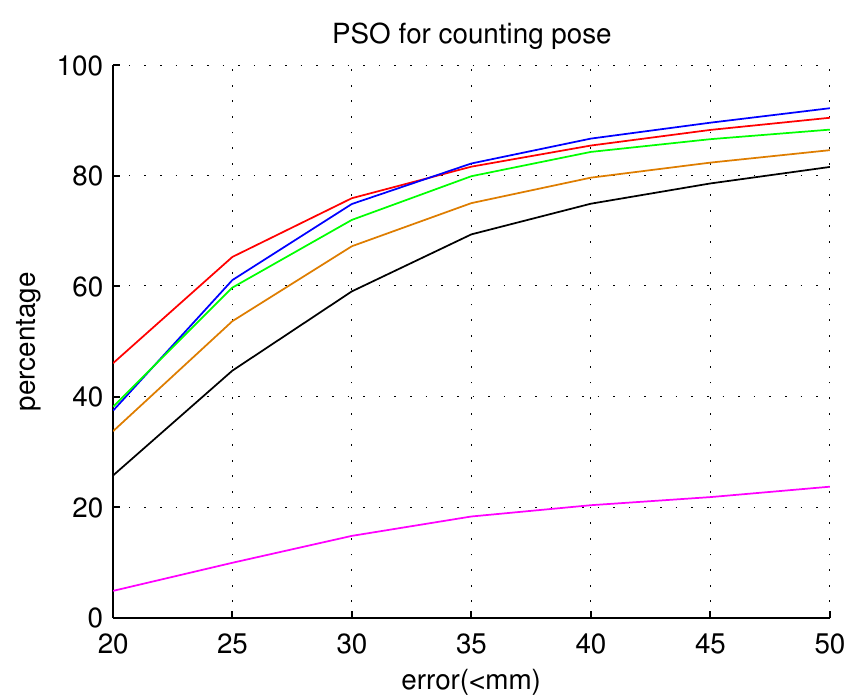} &
        \includegraphics[width=\swthree]{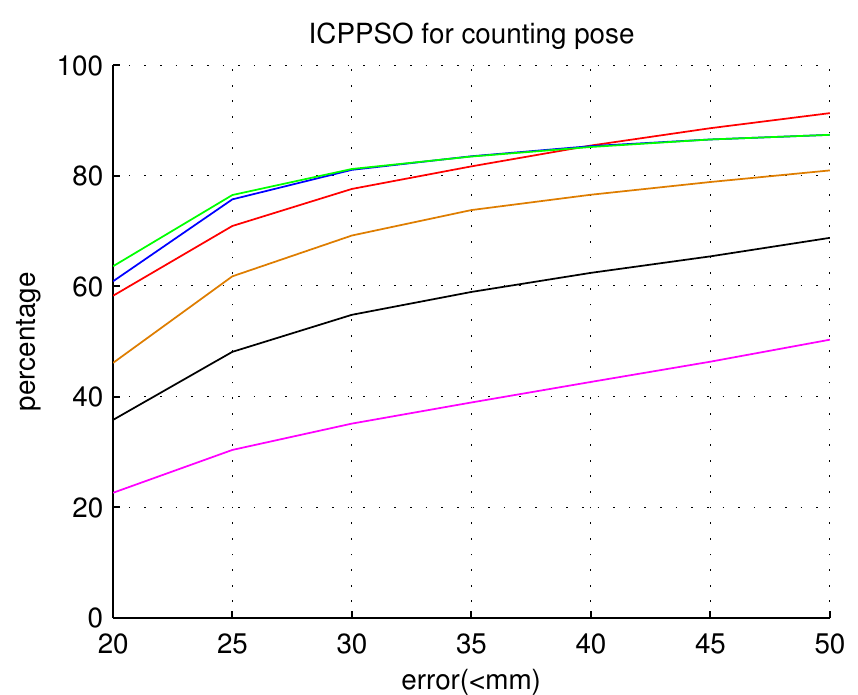} &
        \includegraphics[width=\swthree]{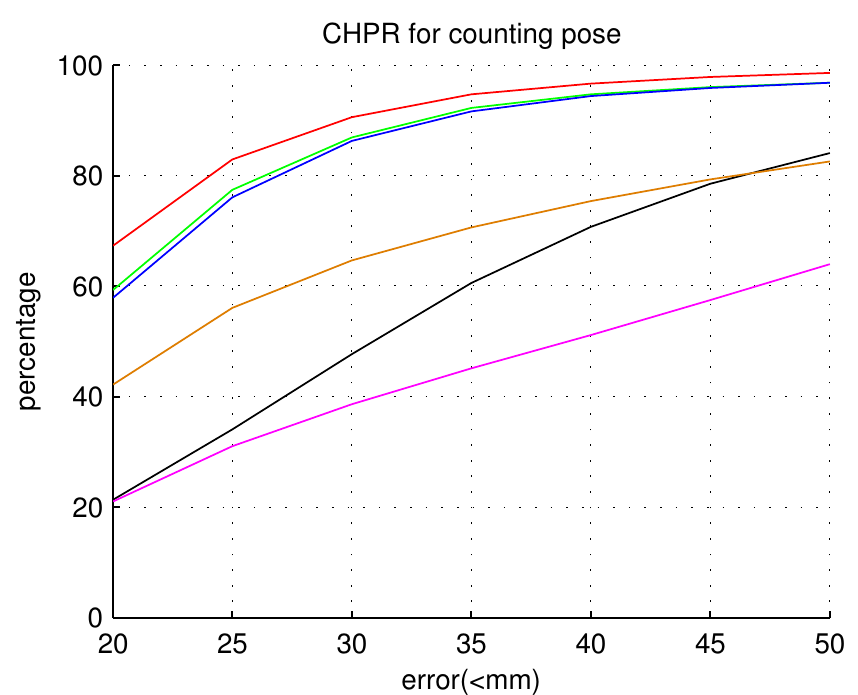} \\
        \includegraphics[width=\swthree]{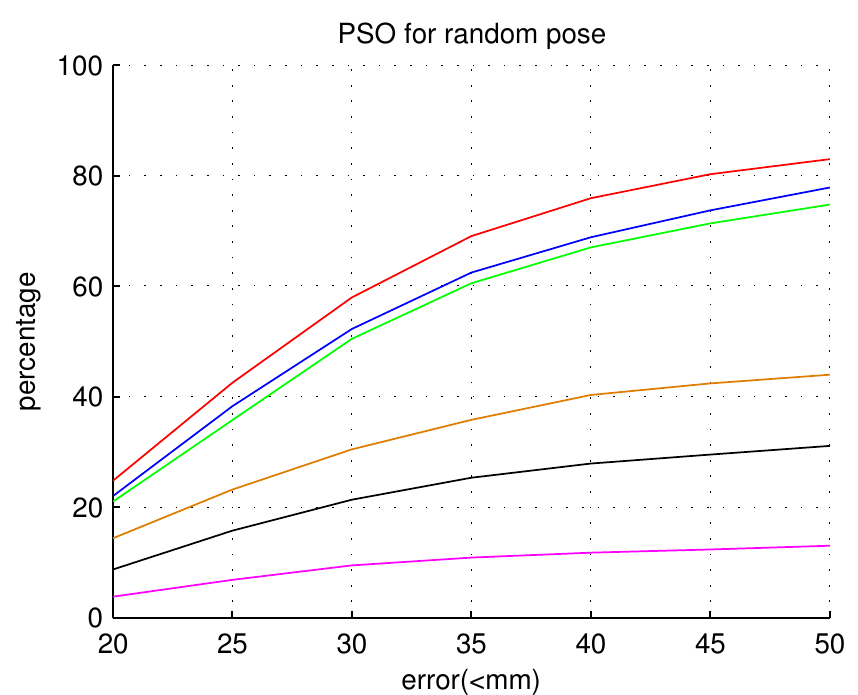} &
        \includegraphics[width=\swthree]{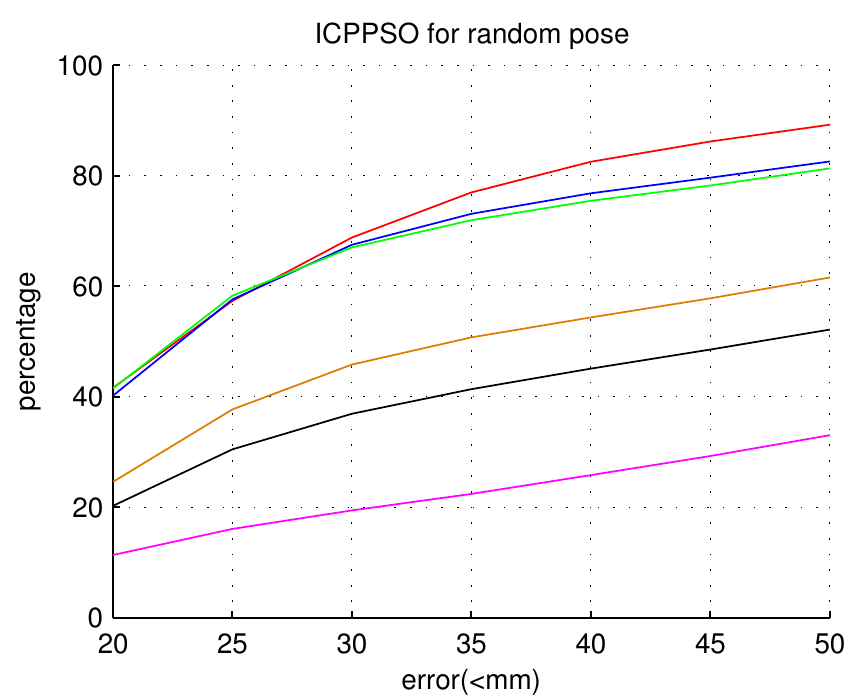} &
        \includegraphics[width=\swthree]{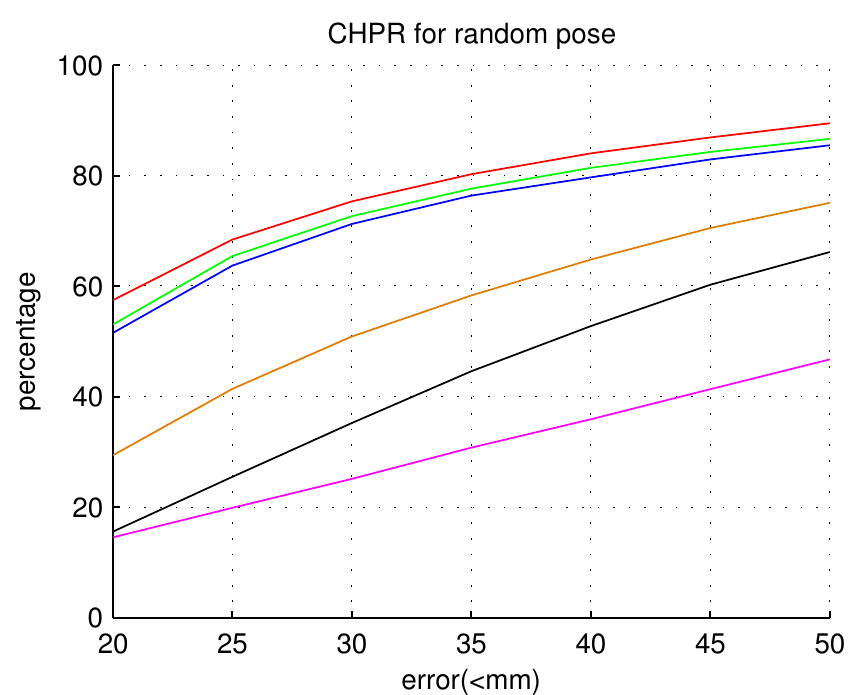} \\
        \small{(a)Tracking: PSO\cite{oikonomidis2011efficient}}&\small{(b)Tracking: ICPPSO\cite{qian2014realtime}}&\small{(c)Estimation: CHPR\cite{sun2015cascaded}}
    \end{tabular}
    \includegraphics[width=0.90\linewidth]{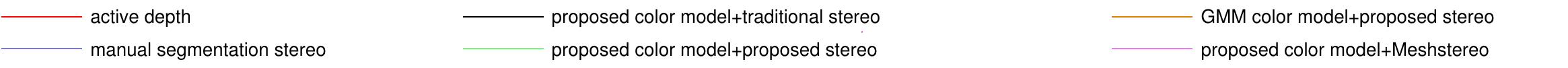} \\
\end{center}
   \caption{\textbf{Average} percentage of all joints that have an maximum error less than x mm (where x starts from$^{\text{\ref{fn:20mm}}}$ 20) under six different backgrounds. The first row presents the tracking/estimation results on simple counting poses while the second row presents difficult random poses. (a)-(c) are the results obtained from two tracking and one estimation algorithms. The red curves are obtained from Intel F200 which is an active depth camera. Note that the performance is the highest on average. The green and blue curves are obtained from the stereo matching and hand segmentation method proposed in Sec. \ref{sec:hand_tracking} and the traditional local stereo method with manual hand segmentation results. Note that the green and blue curves are very close to each other and are comparable to the red curves (which represent the performance of an active depth sensor). The brown curves are obtained from proposed stereo matching method with generic skin color model \cite{jones2002statistical}. Brown curves do not consider \textit{B4} and \textit{B6} since the generic skin color model completely fails on these video sequences. The pink curves are from Meshstereo \cite{zhang2015meshstereo} with proposed color model. \textit{B2} is not considered in Meshstereo as the quality of the estimated hand disparities is too low for the adopted hand tracking/estimation algorithms. See text for details. Evaluation on individual background is presented in the \reffig{fig:individual_result}.
   %for comparison of Intel F200 active depth camera and stereo matching with counting pose and random pose tracked by PSO \cite{oikonomidis2011efficient}, ICPPSO \cite{qian2014realtime} and CHPR \cite{sun2015cascaded}. All the stereo matchings use guided filter for cost aggregation and we test different stereo matchings with manually segmented hand mask, traditional stereo and proposed stereo. For proposed stereo, we get the color model from both a generic GMM skin color model \cite{jones2002statistical} and our proposed training based model. In \textit{B4} and \textit{B6}, the tracking of GMM with our proposed stereo are not plotted since the model in \cite{jones2002statistical} is too bad under this two scenarios (forth and sixth row of \reffig{fig:mask_comp} (b)).
   }
\label{fig:all_result}
\end{figure*}

\renewcommand{\tabcolsep}{0.1 pt}
\begin{figure*}[!ht]
\begin{center}
    \begin{tabular}{cccccc}
        \includegraphics[width=\swseven]{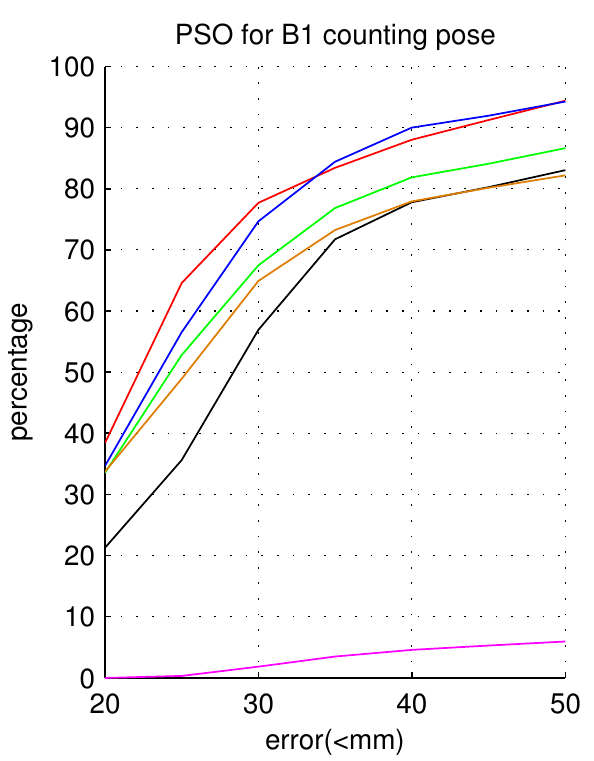} &
        \includegraphics[width=\swseven]{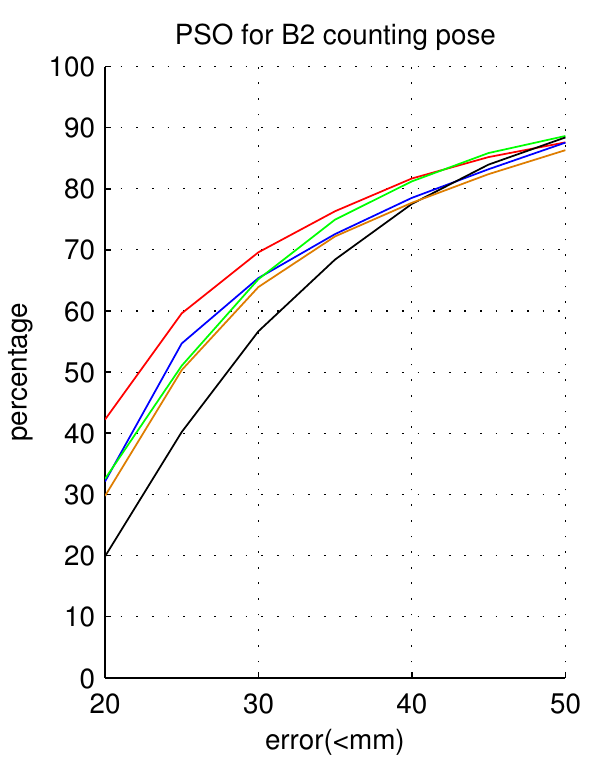} &
        \includegraphics[width=\swseven]{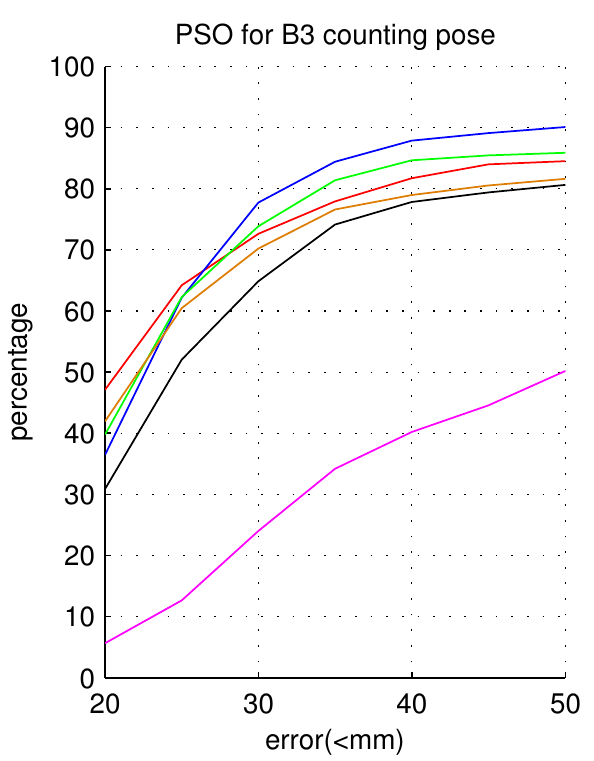} &
        \includegraphics[width=\swseven]{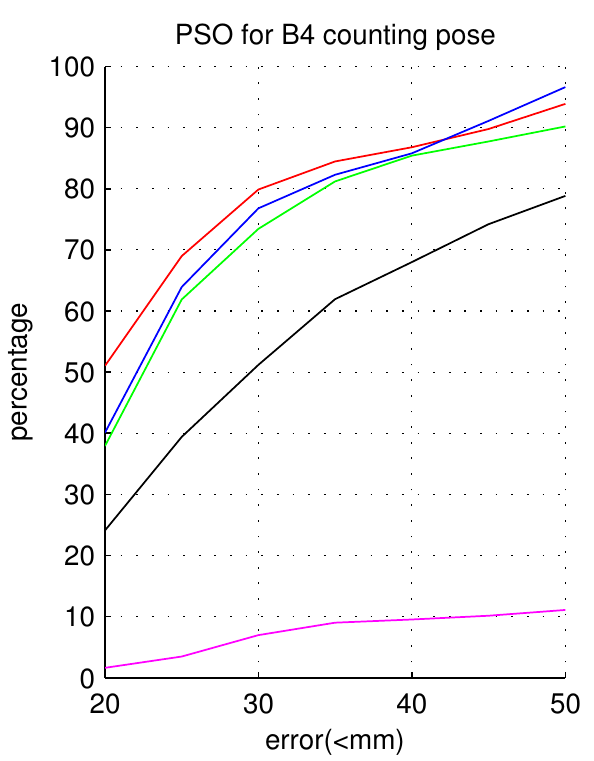} &
        \includegraphics[width=\swseven]{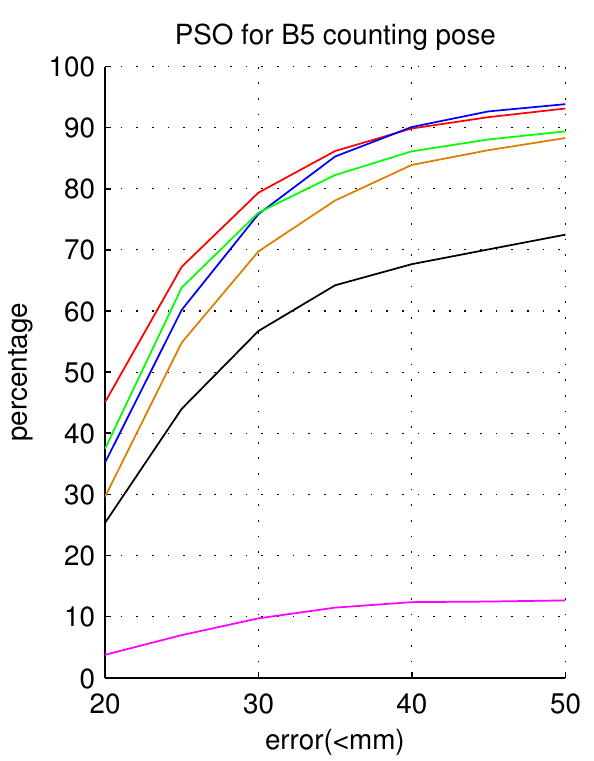} &
        \includegraphics[width=\swseven]{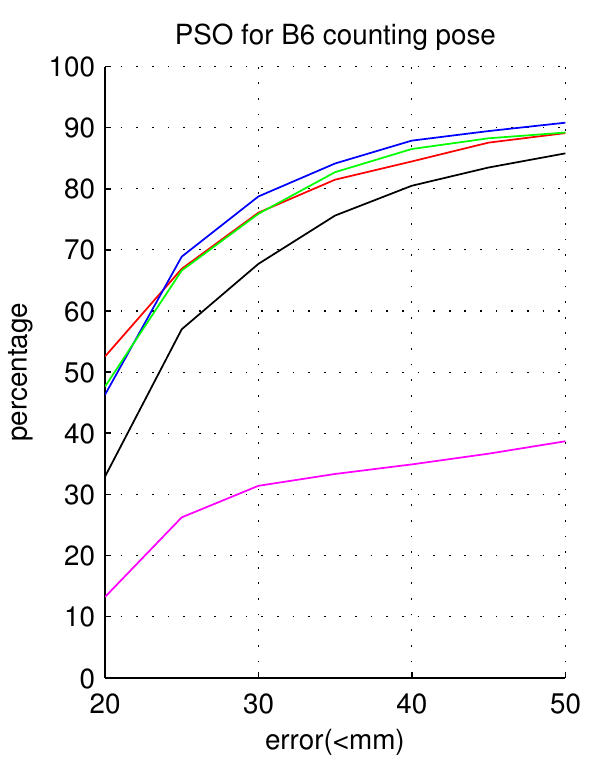} \\
        \includegraphics[width=\swseven]{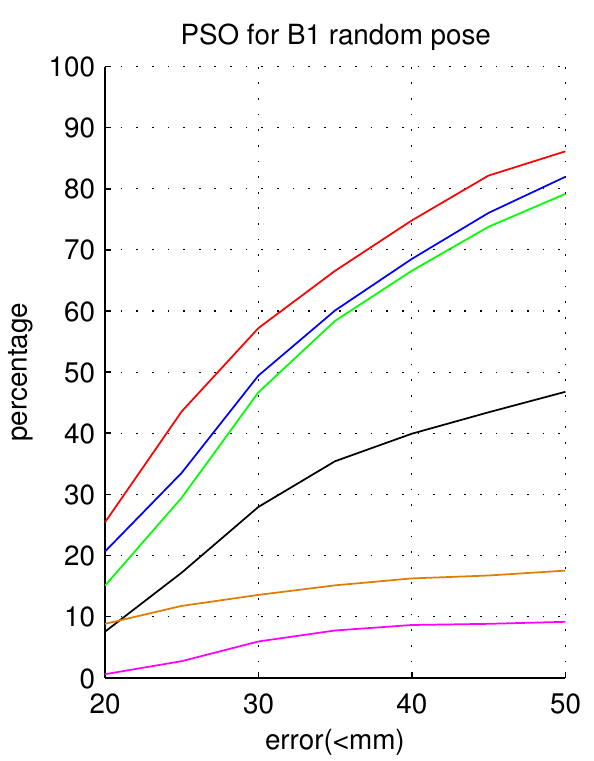} &
        \includegraphics[width=\swseven]{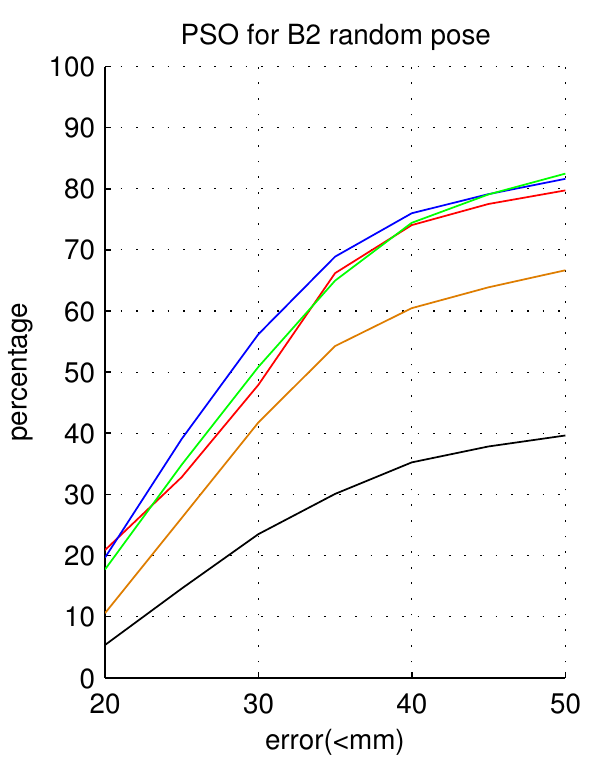} &
        \includegraphics[width=\swseven]{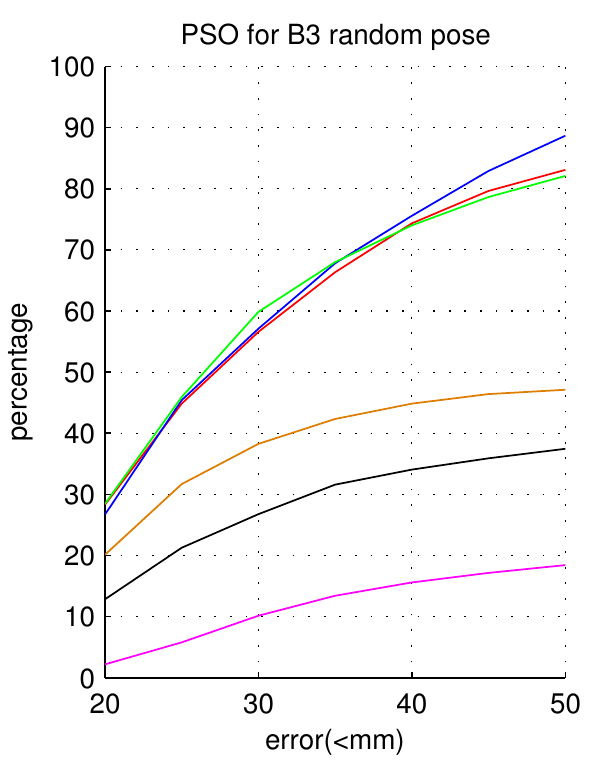} &
        \includegraphics[width=\swseven]{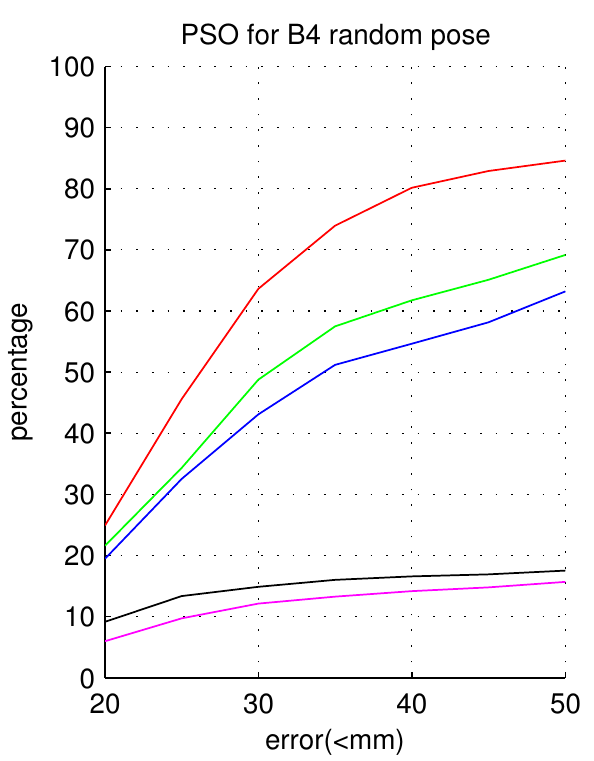} &
        \includegraphics[width=\swseven]{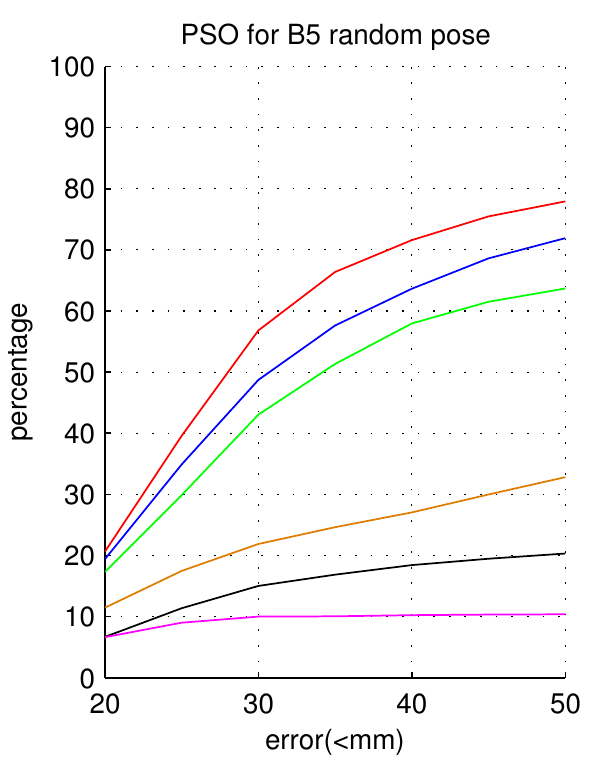} &
        \includegraphics[width=\swseven]{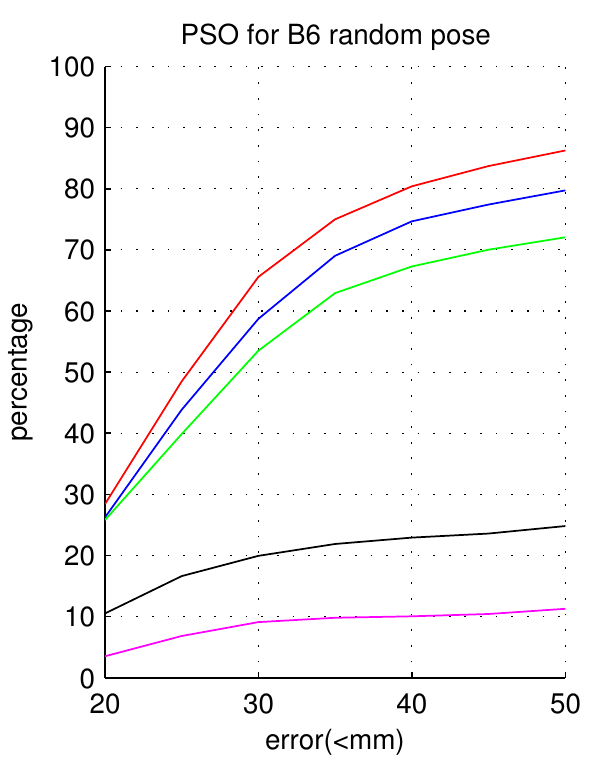} \\
        \includegraphics[width=\swseven]{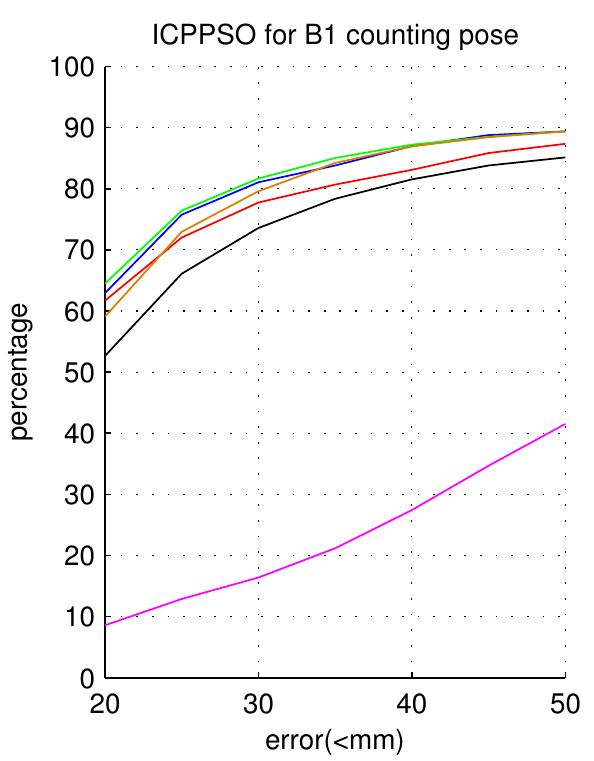} &
        \includegraphics[width=\swseven]{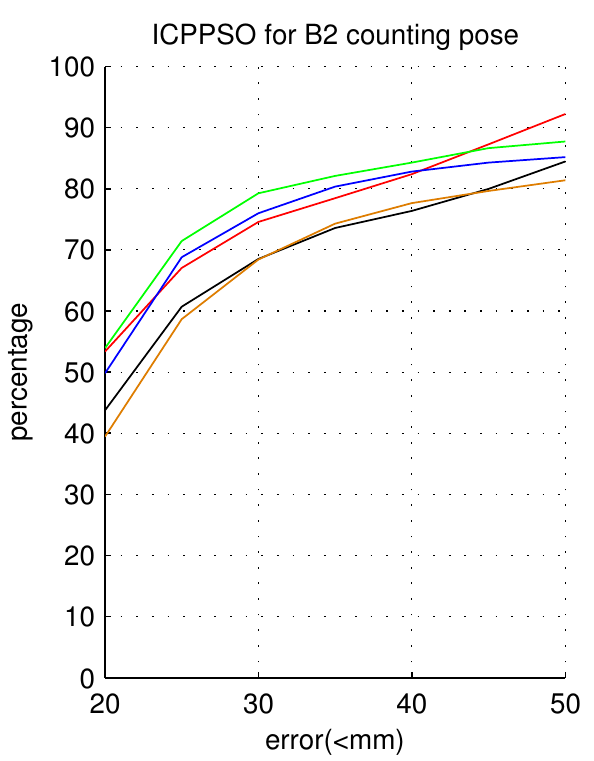} &
        \includegraphics[width=\swseven]{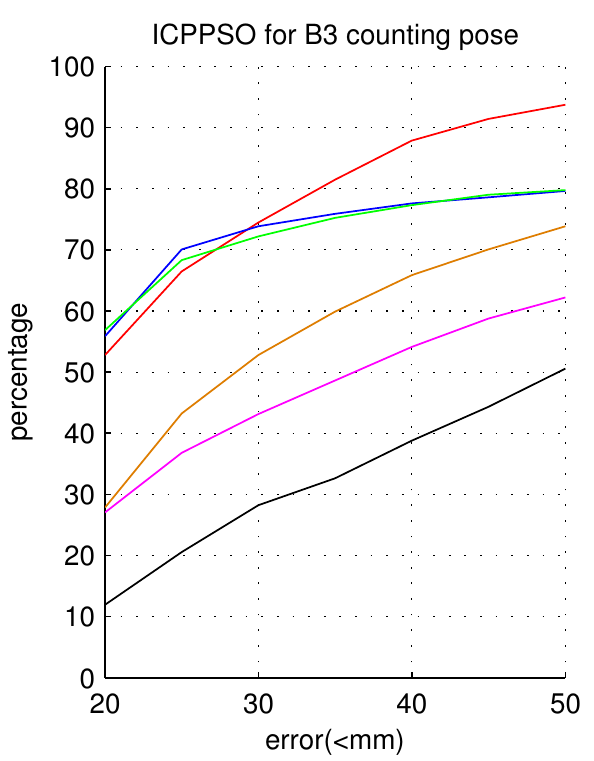} &
        \includegraphics[width=\swseven]{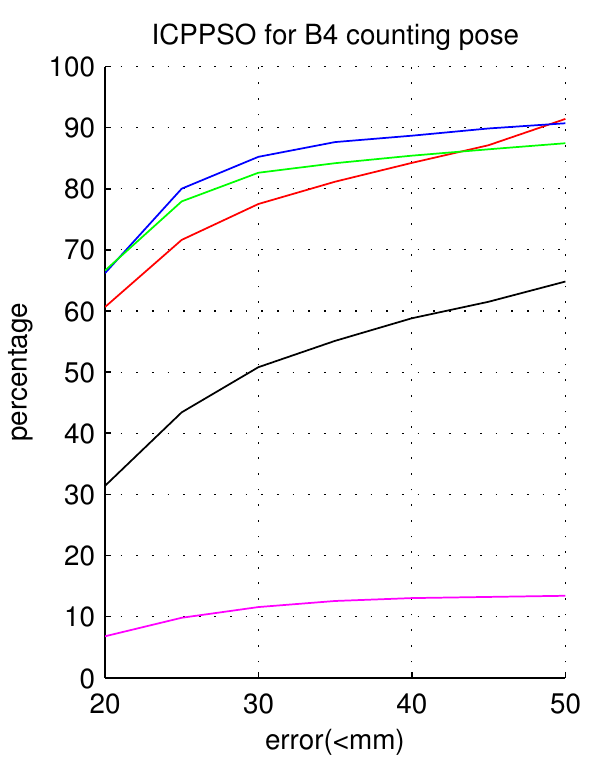} &
        \includegraphics[width=\swseven]{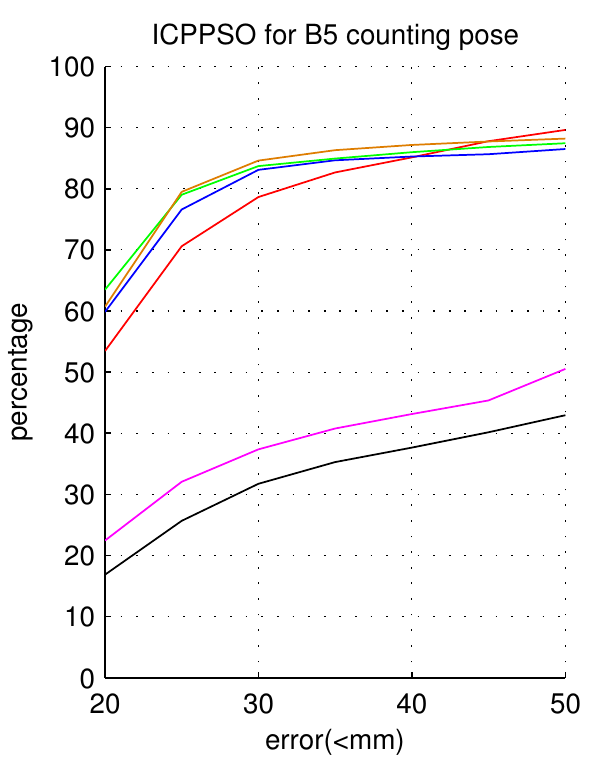} &
        \includegraphics[width=\swseven]{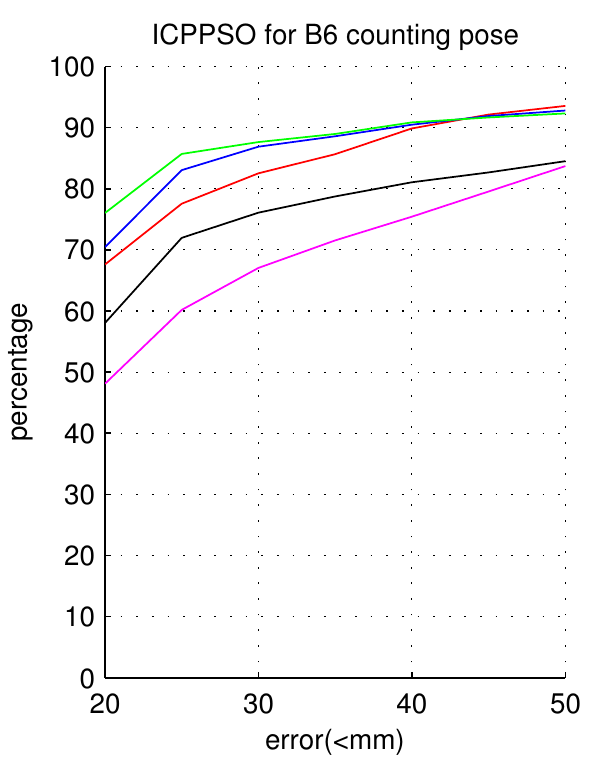} \\
        \includegraphics[width=\swseven]{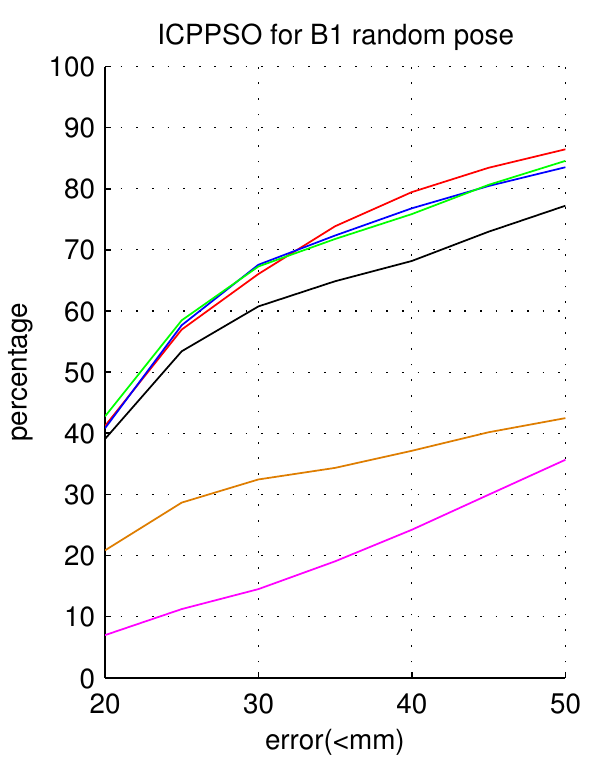} &
        \includegraphics[width=\swseven]{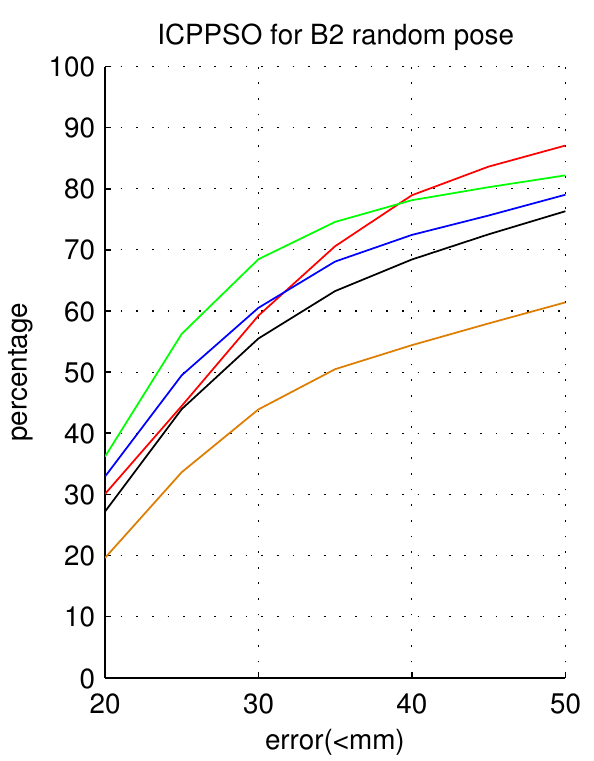} &
        \includegraphics[width=\swseven]{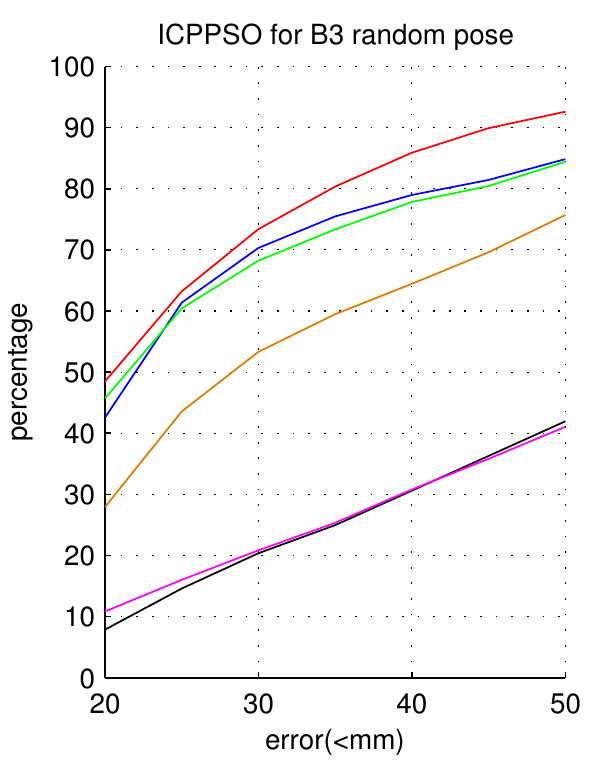} &
        \includegraphics[width=\swseven]{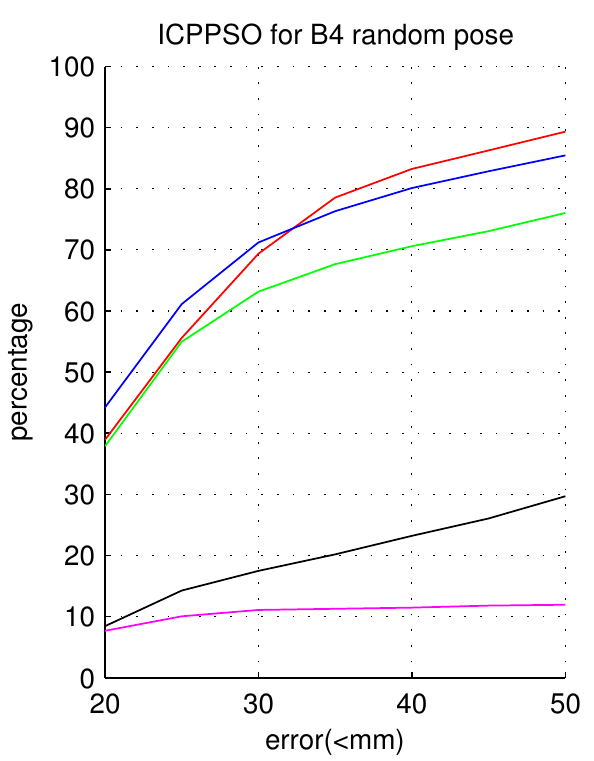} &
        \includegraphics[width=\swseven]{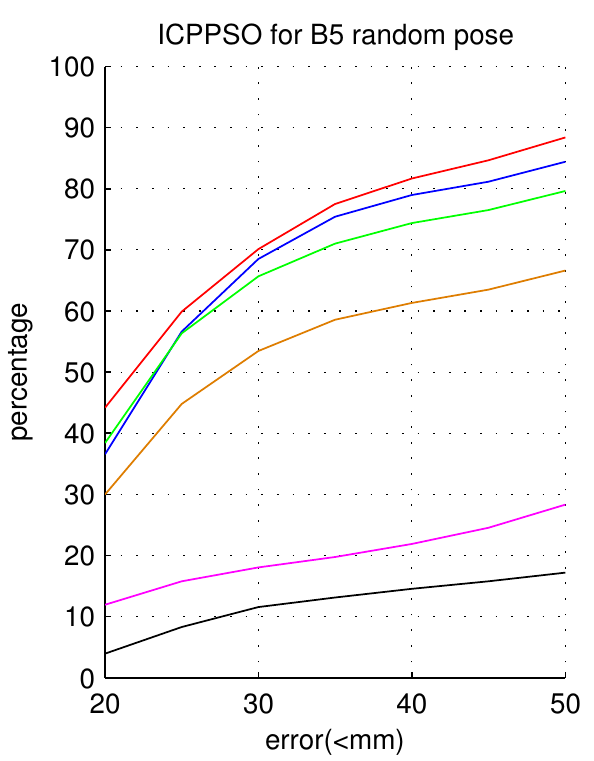} &
        \includegraphics[width=\swseven]{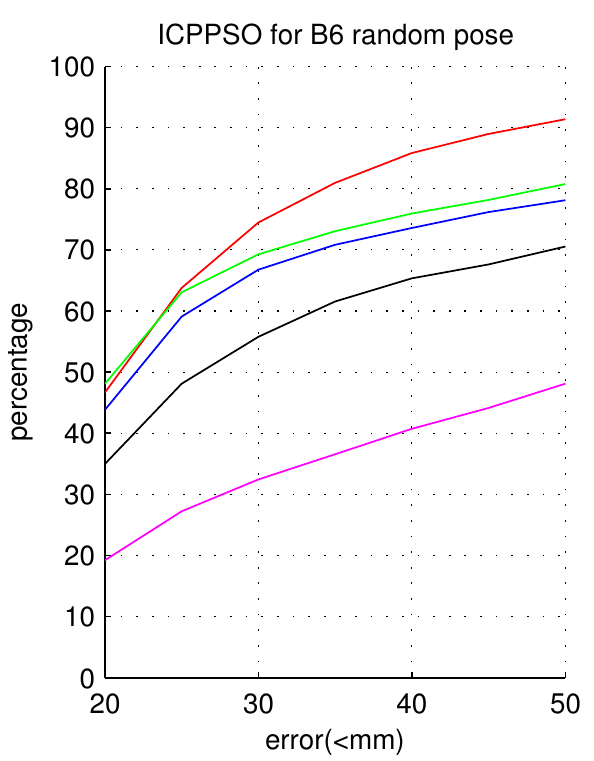} \\
        \includegraphics[width=\swseven]{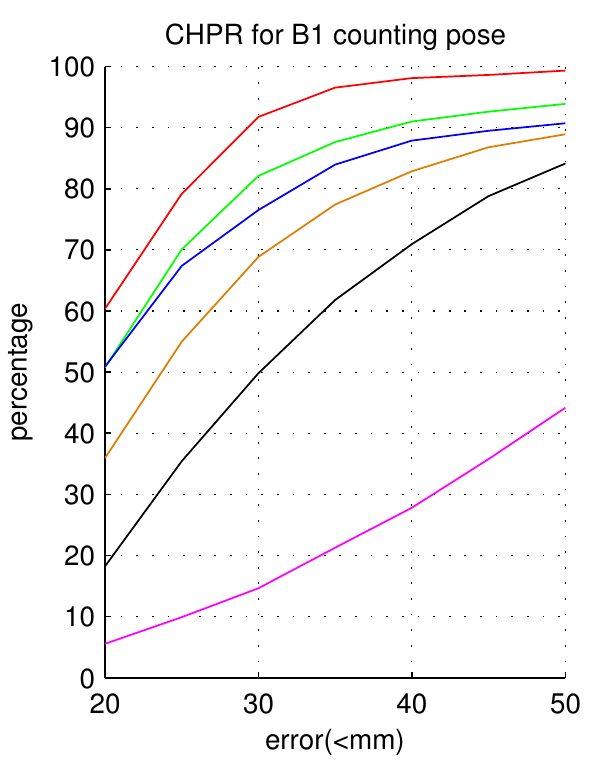} &
        \includegraphics[width=\swseven]{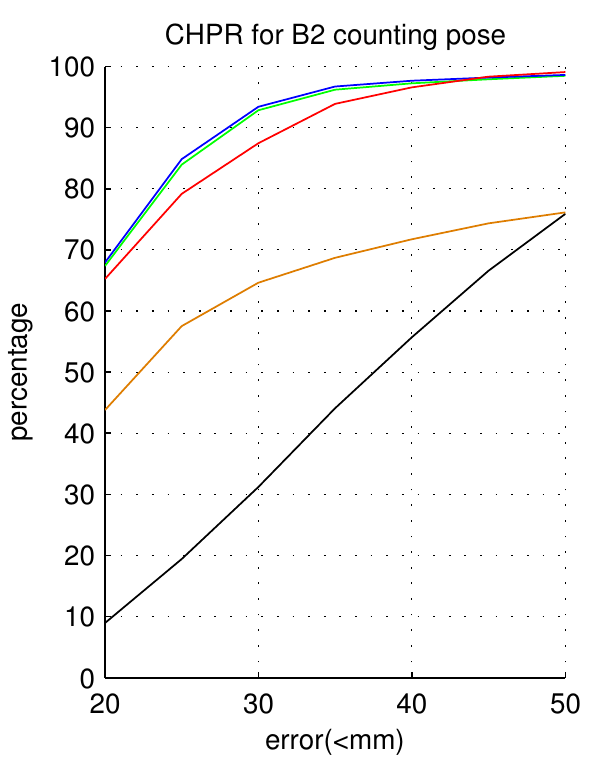} &
        \includegraphics[width=\swseven]{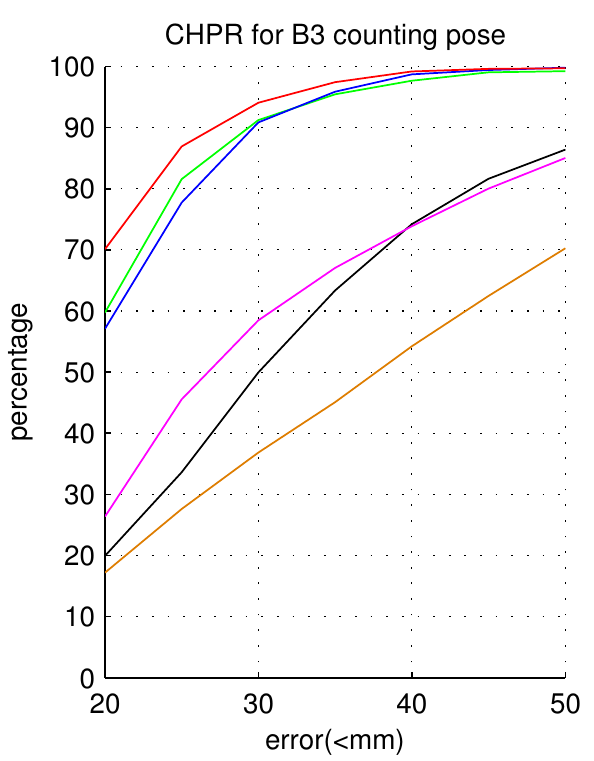} &
        \includegraphics[width=\swseven]{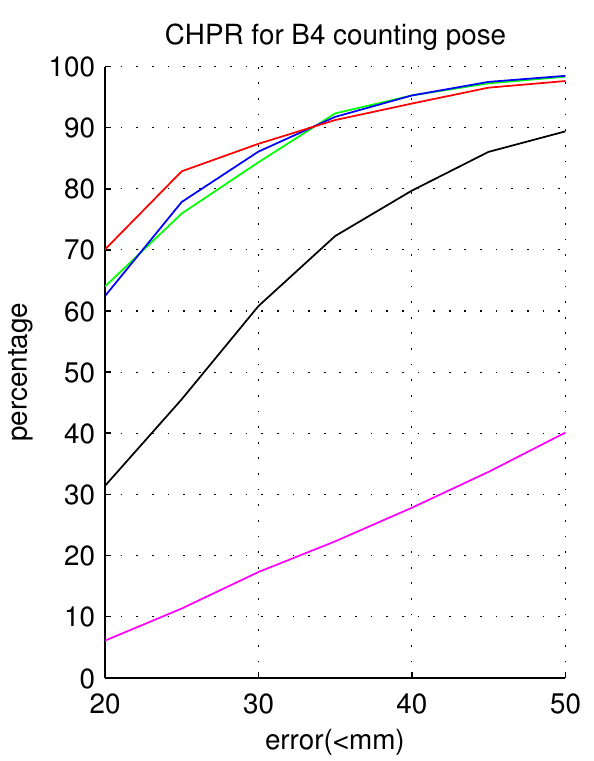} &
        \includegraphics[width=\swseven]{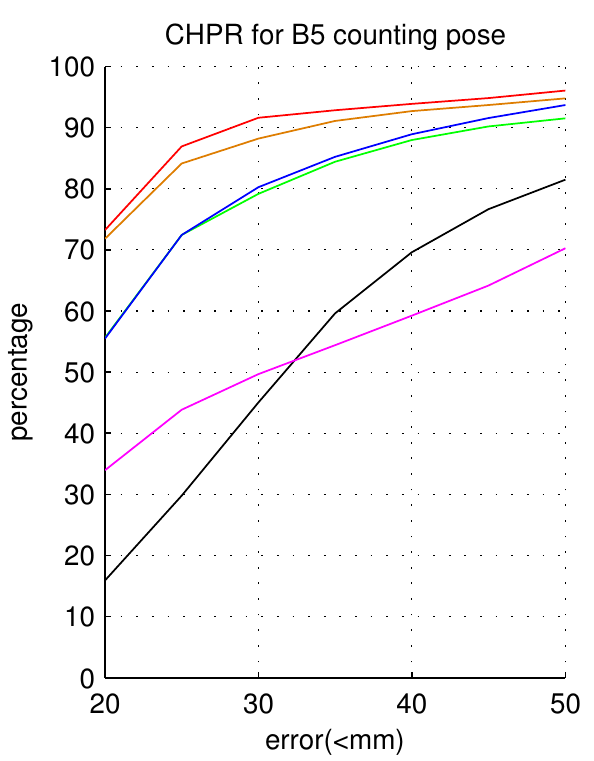} &
        \includegraphics[width=\swseven]{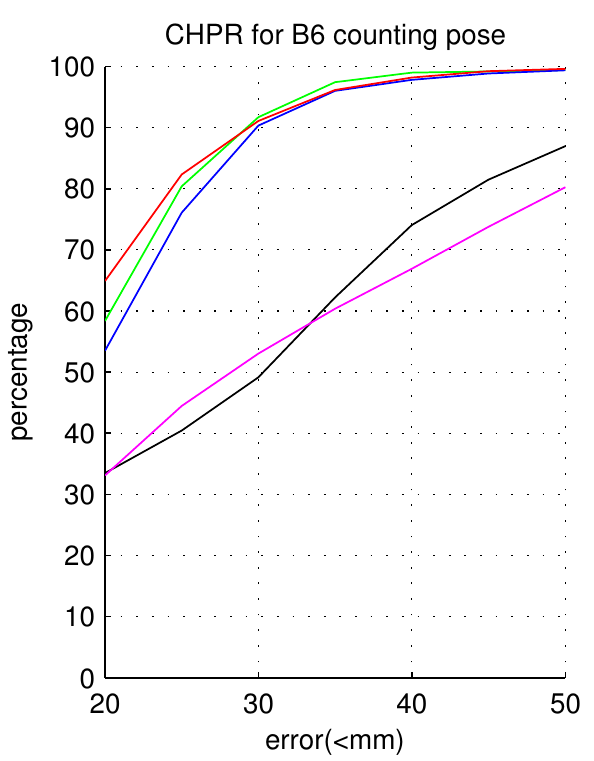} \\
        \includegraphics[width=\swseven]{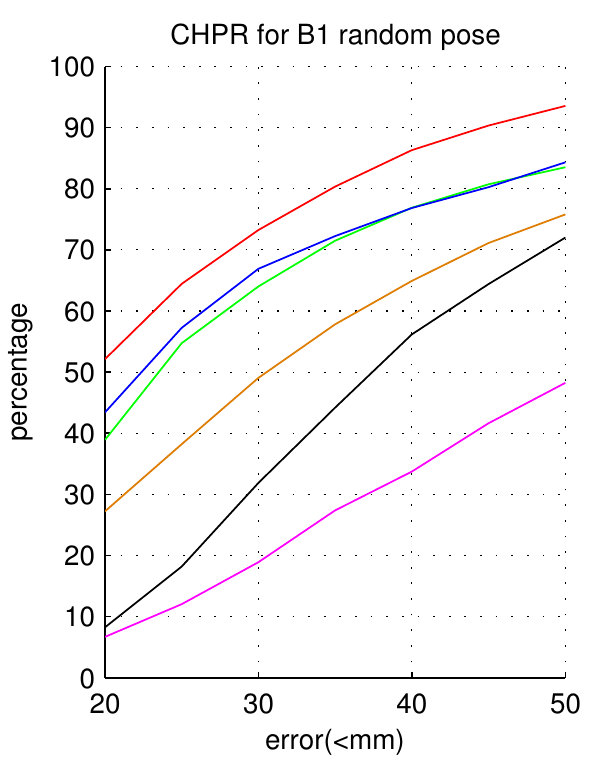} &
        \includegraphics[width=\swseven]{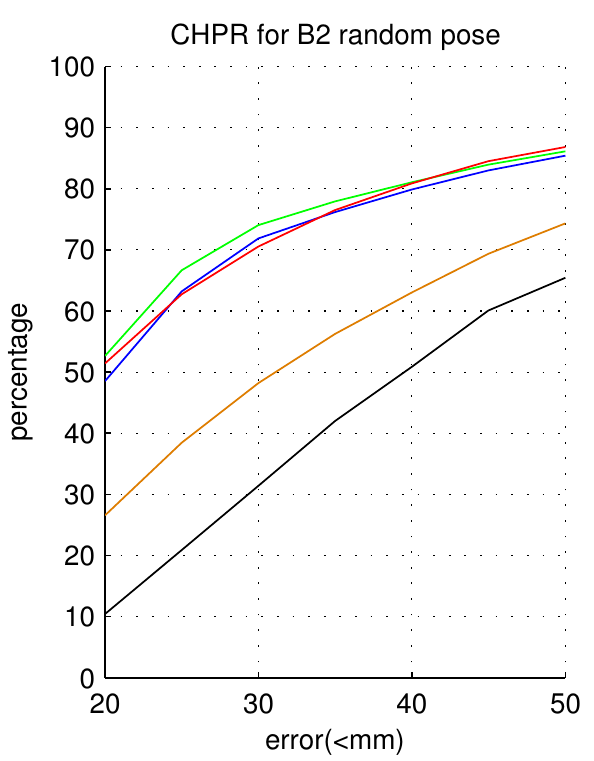} &
        \includegraphics[width=\swseven]{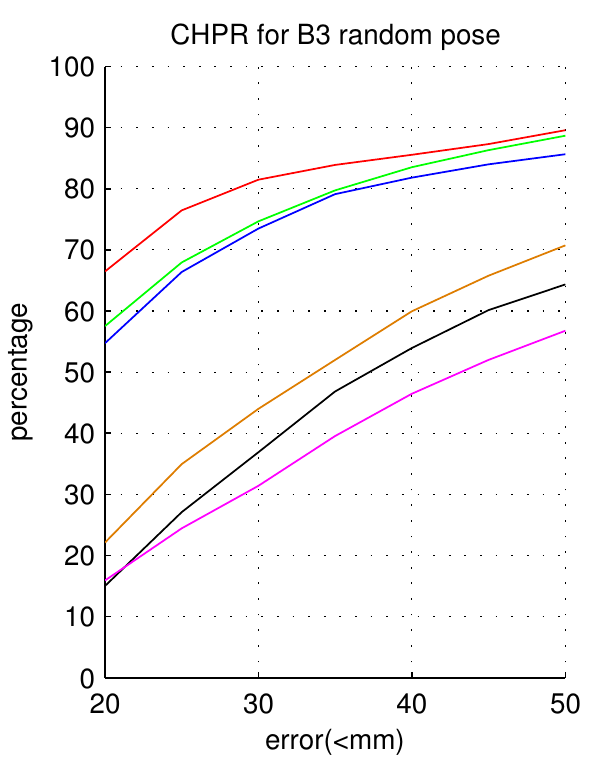} &
        \includegraphics[width=\swseven]{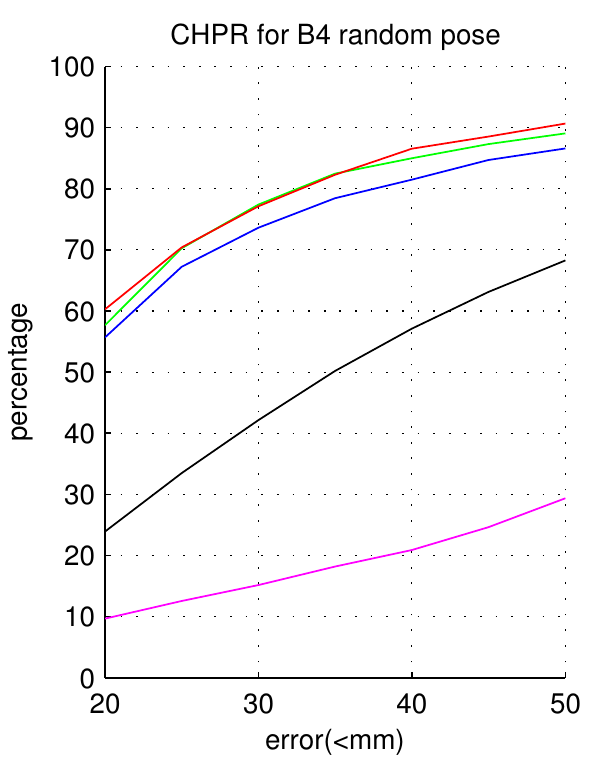} &
        \includegraphics[width=\swseven]{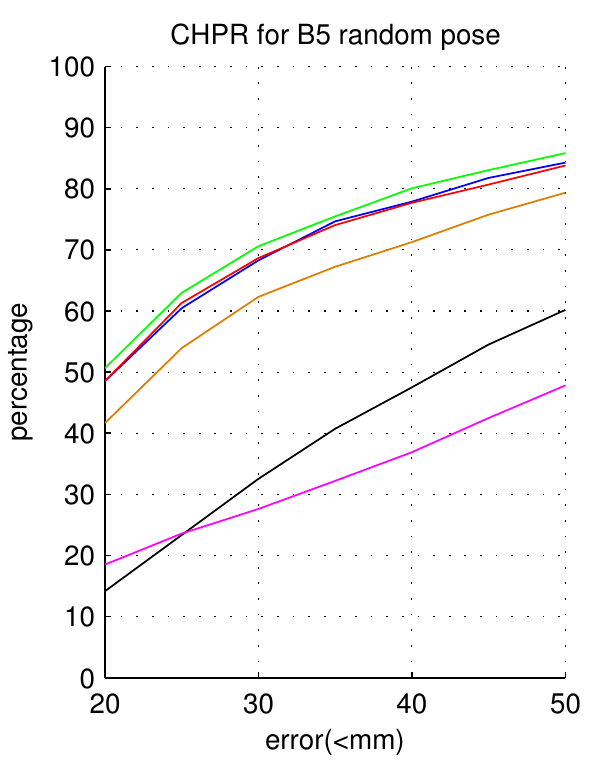} &
        \includegraphics[width=\swseven]{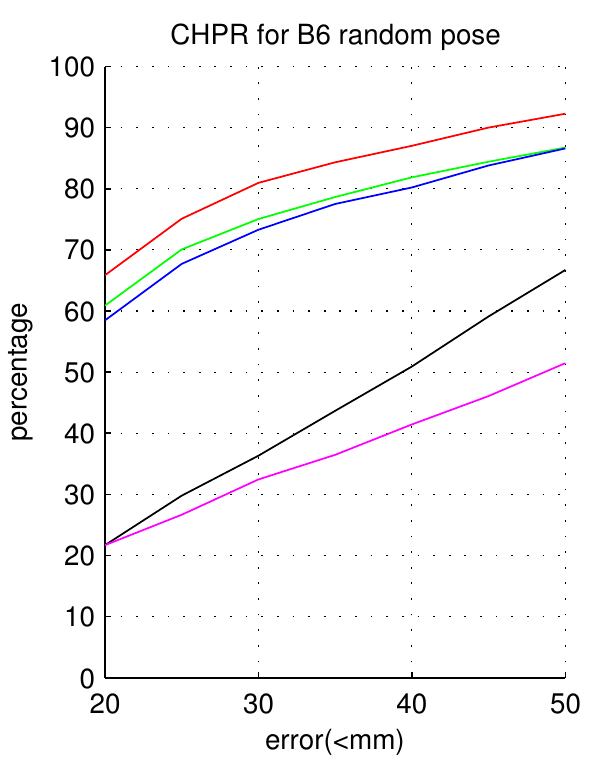} \\
    \end{tabular}
    \includegraphics[width=0.90\linewidth]{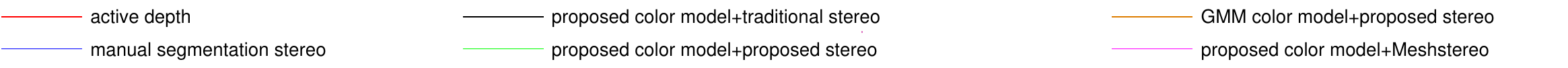} \\
\end{center}
   \caption{Percentage of all joints that have an maximum error less than x mm under different scenarios by \textbf{PSO} \cite{oikonomidis2011efficient}, \textbf{ICPPSO} \cite{qian2014realtime} and \textbf{CHPR} \cite{sun2015cascaded}. The red curves are obtained from Intel F200 which is an active depth camera. The green and blue curves are obtained from the stereo matching with hand segmentation method proposed in Sec. \ref{sec:hand_tracking} and the traditional local stereo method with manual hand segmentation. The dark curves are obtained from the traditional local stereo with the proposed hand segmentation. The brown curves are obtained from proposed stereo matching method with generic skin color model \cite{jones2002statistical}. Brown curves are not plotted for \textit{B4} and \textit{B6} since the generic skin color model completely fails on these video sequences. The pink curves are from Meshstereo \cite{zhang2015meshstereo} with proposed color model. Meshstereo was ignored in \textit{B2} as the quality of the estimated hand disparities is too low for the adopted hand tracking/estimation algorithms.
   }
\label{fig:individual_result}
\end{figure*}

\renewcommand{\tabcolsep}{0.1 pt}
\begin{figure*}[!ht]
\def\swvisualmodels{0.15\linewidth}
\begin{center}
    \begin{tabular}{cccccccc}
        \rotatebox{90}{(a)Color}&&
        \includegraphics[width=\swvisualmodels]{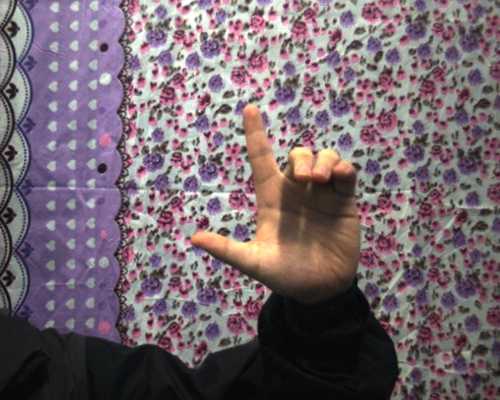} &
        \includegraphics[width=\swvisualmodels]{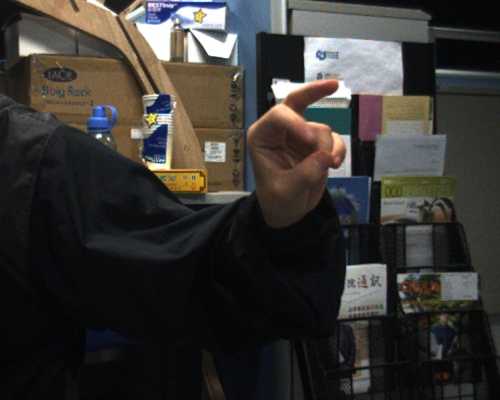} &
        \includegraphics[width=\swvisualmodels]{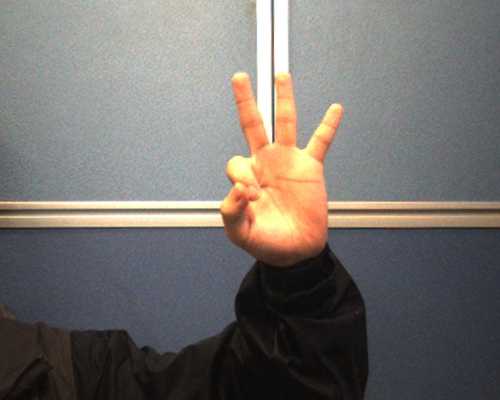} &
        \includegraphics[width=\swvisualmodels]{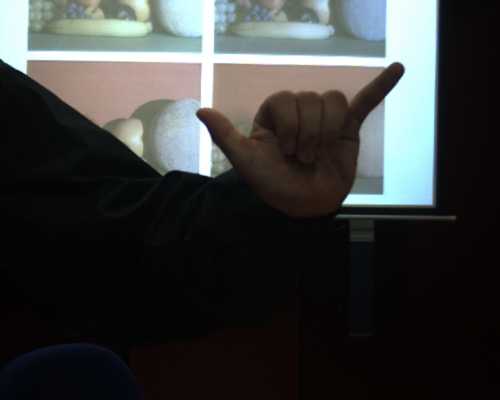} &
        \includegraphics[width=\swvisualmodels]{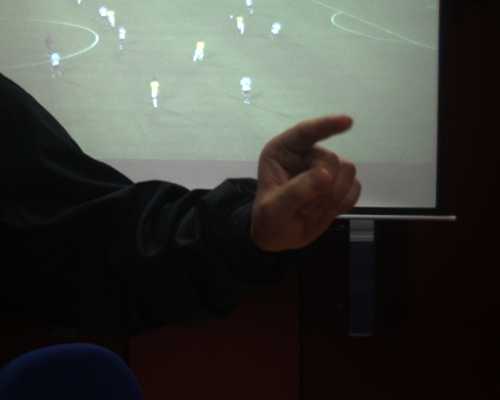} &
        \includegraphics[width=\swvisualmodels]{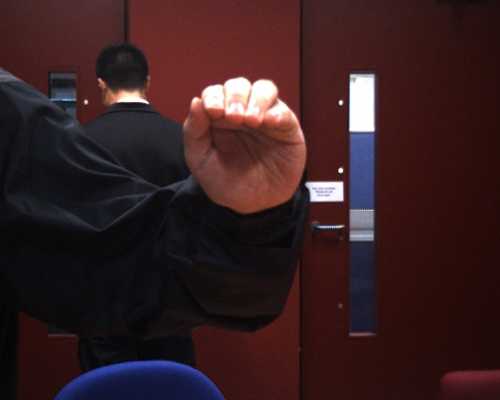} \\
        \rotatebox{90}{(b)Hand}&\rotatebox{90}{disparities}&
        \includegraphics[width=\swvisualmodels]{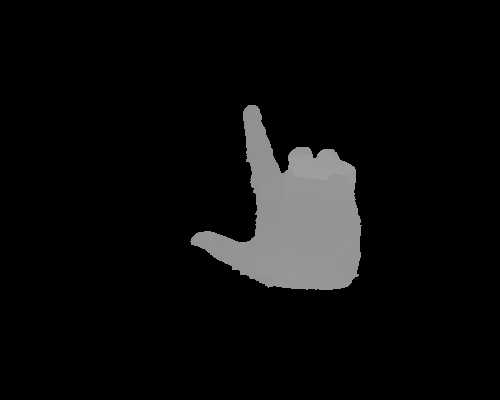} &
        \includegraphics[width=\swvisualmodels]{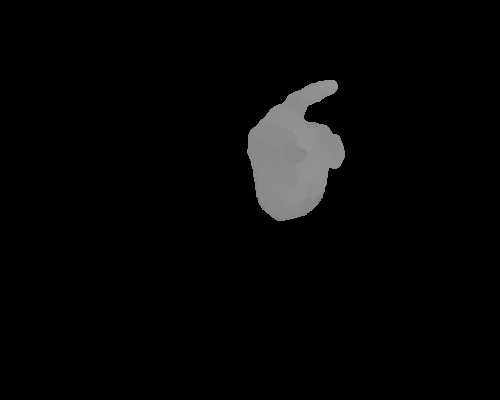} &
        \includegraphics[width=\swvisualmodels]{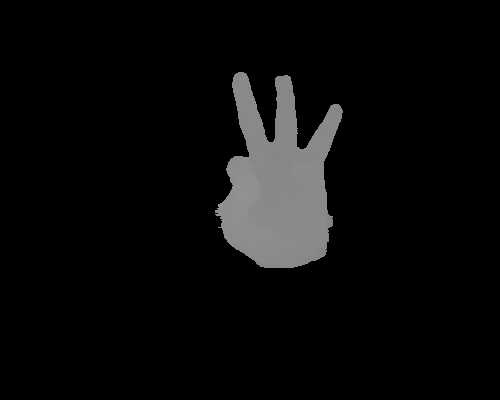} &
        \includegraphics[width=\swvisualmodels]{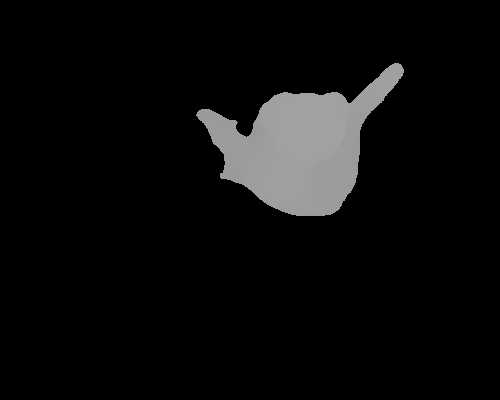} &
        \includegraphics[width=\swvisualmodels]{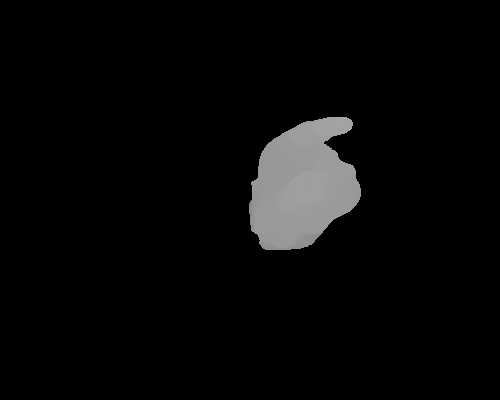} &
        \includegraphics[width=\swvisualmodels]{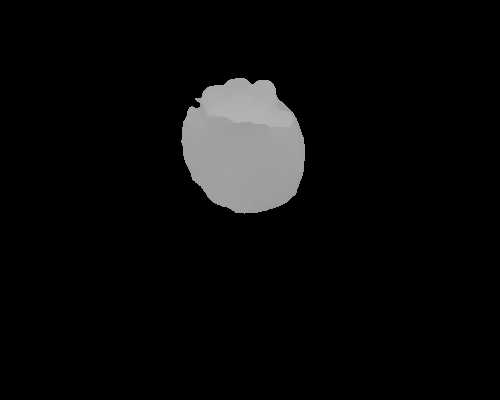} \\
        \rotatebox{90}{(c)\cite{oikonomidis2011efficient}}&\rotatebox{90}{PSO}&
        \includegraphics[width=\swvisualmodels]{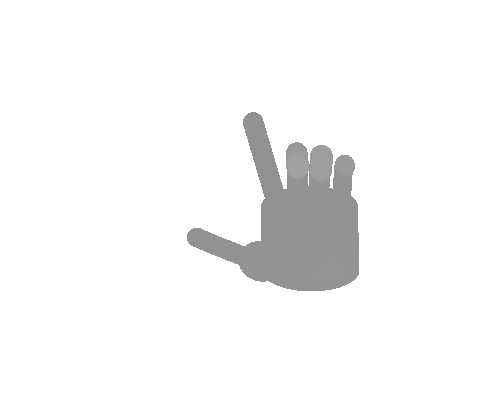} &
        \includegraphics[width=\swvisualmodels]{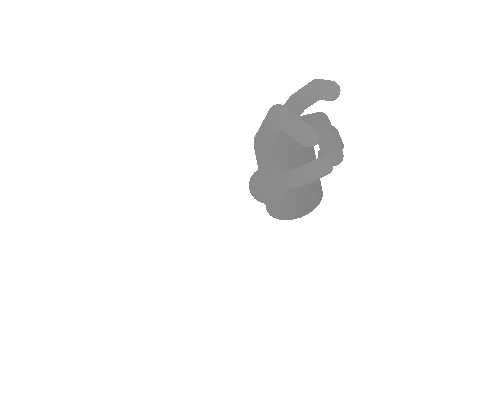} &
        \includegraphics[width=\swvisualmodels]{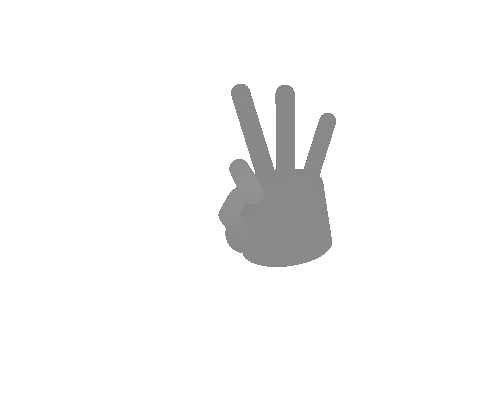} &
        \includegraphics[width=\swvisualmodels]{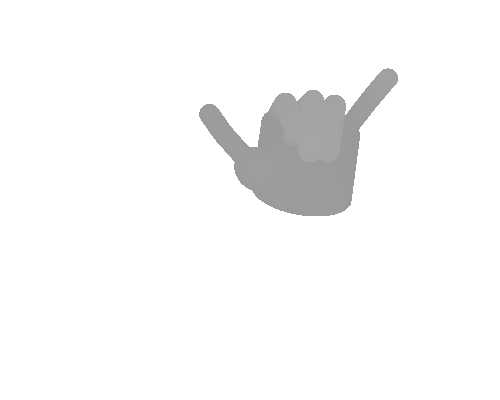} &
        \includegraphics[width=\swvisualmodels]{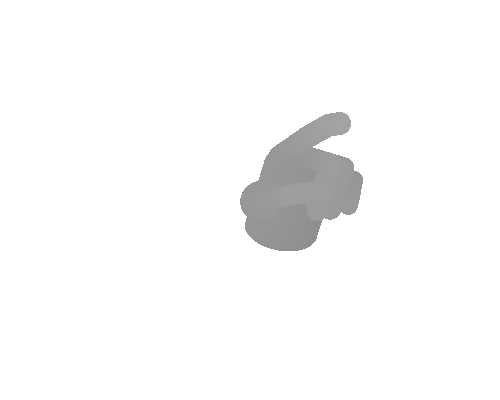} &
        \includegraphics[width=\swvisualmodels]{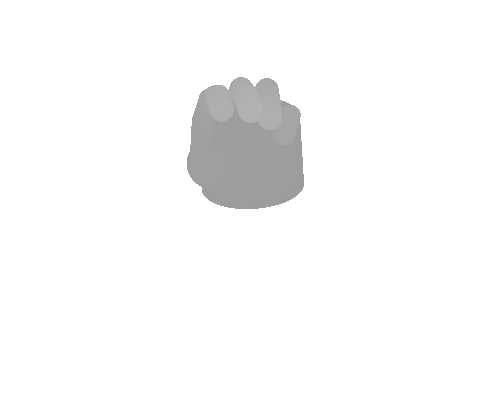} \\
        \rotatebox{90}{(d)\cite{qian2014realtime}}&\rotatebox{90}{ICPPSO}&
        \includegraphics[width=\swvisualmodels]{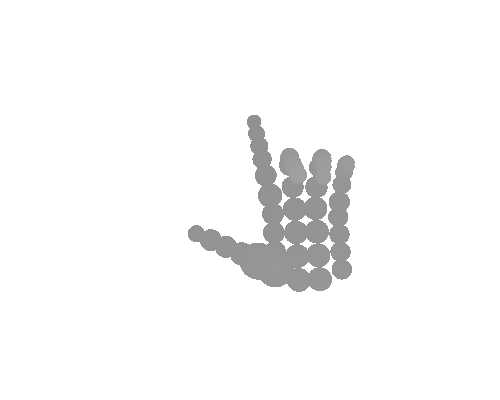} &
        \includegraphics[width=\swvisualmodels]{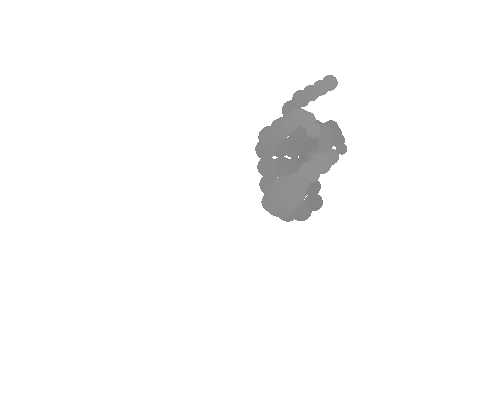} &
        \includegraphics[width=\swvisualmodels]{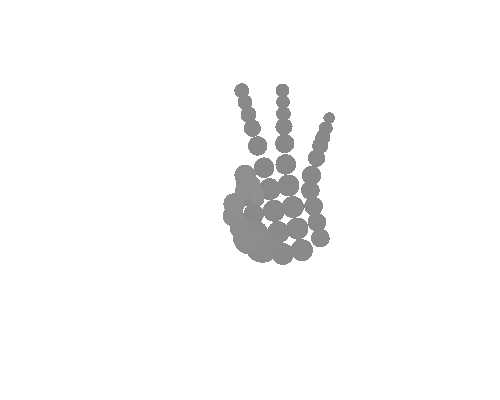} &
        \includegraphics[width=\swvisualmodels]{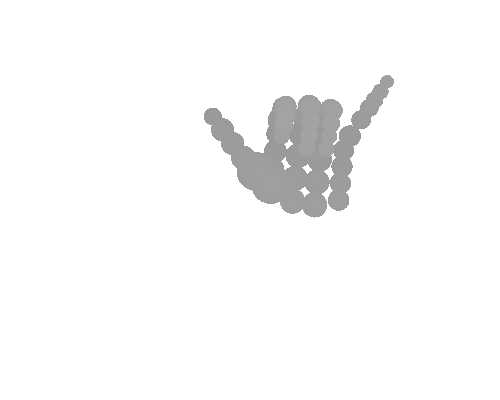} &
        \includegraphics[width=\swvisualmodels]{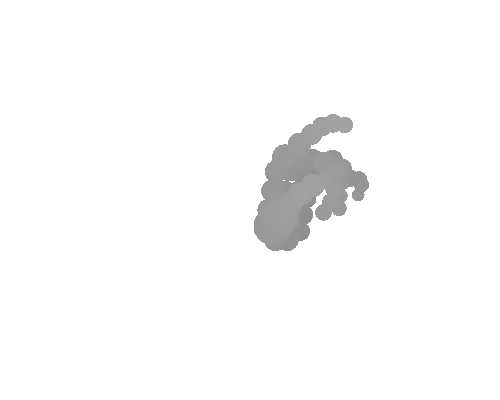} &
        \includegraphics[width=\swvisualmodels]{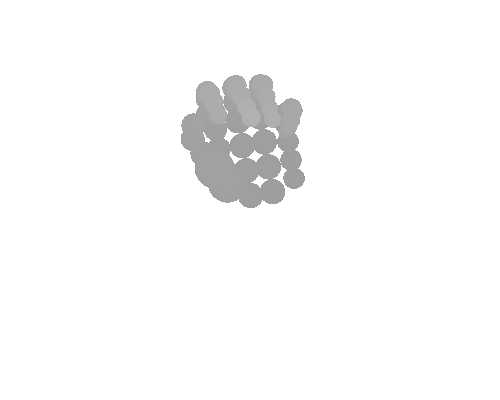} \\
        \rotatebox{90}{(e)\cite{sun2015cascaded}}&\rotatebox{90}{CHPR}&
        \includegraphics[width=\swvisualmodels]{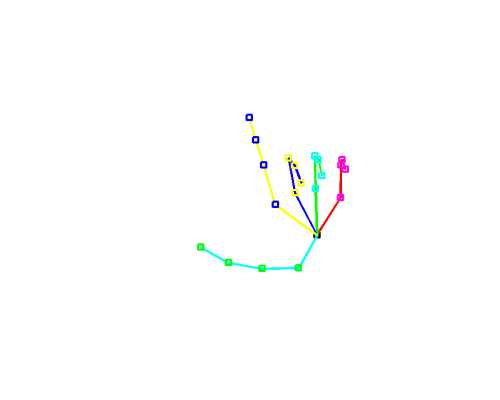} &
        \includegraphics[width=\swvisualmodels]{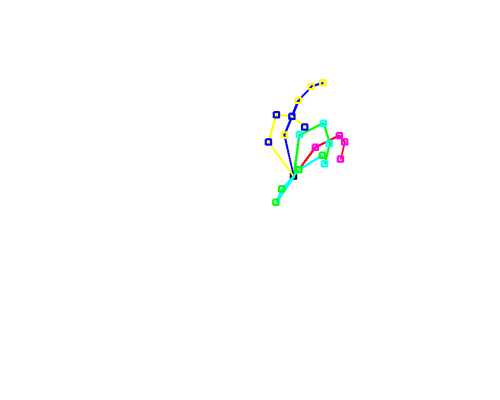} &
        \includegraphics[width=\swvisualmodels]{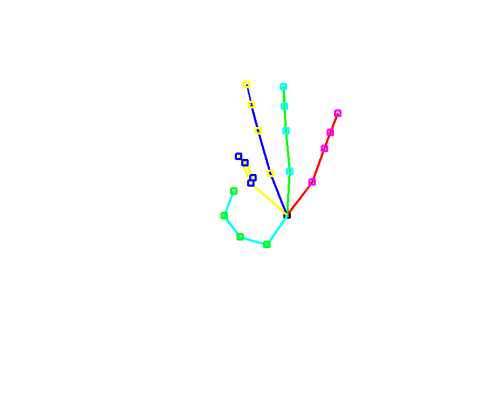} &
        \includegraphics[width=\swvisualmodels]{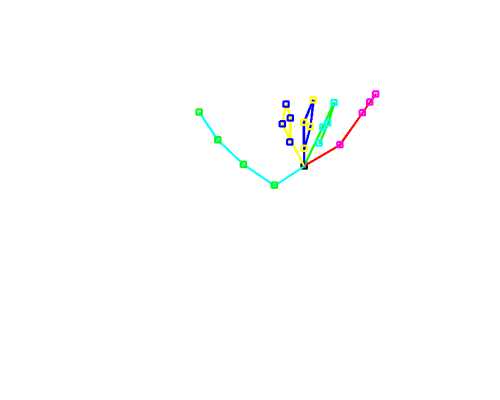} &
        \includegraphics[width=\swvisualmodels]{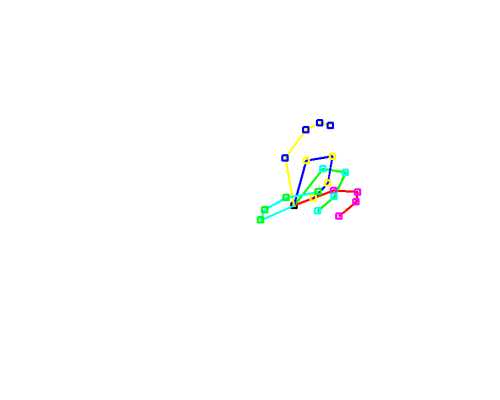} &
        \includegraphics[width=\swvisualmodels]{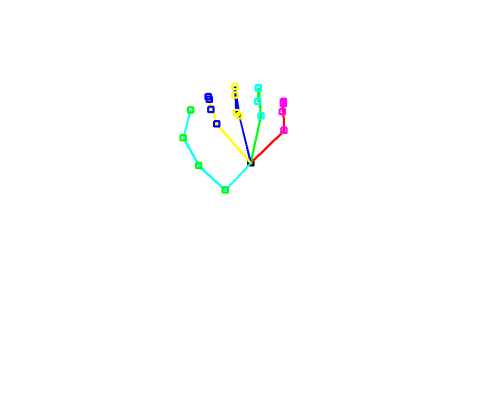} \\
        &&\textit{B1}&\textit{B2}&\textit{B3}&\textit{B4}&\textit{B5}&\textit{B6}\\
    \end{tabular}
\end{center}
   \caption{(c)-(e) present hand pose models estimated from the proposed method with different tracking/estimation algorithms\cite{oikonomidis2011efficient,qian2014realtime,sun2015cascaded}, respectively. Note that they are visually consistent with the segmented hands in (b) and the color images in (a) under different backgrounds.
   }
\label{fig:est_result}
\end{figure*}

This section presents quantitative and visual comparison on hand pose tracking and estimation using both active depth camera and the proposed passive stereo system. It demonstrates that passive stereo can achieve tracking/estimation performance comparable to the active depth camera on challenging real-world scenarios.

The stereo matching algorithm proposed in Sec. \ref{sec:stereo} is indeed a local stereo method that uses Census transform for matching cost computation and the guided image filter for cost aggregation. As a result, it is also compared with the traditional local stereo algorithm that adopts the same matching cost and cost aggregation method.
Besides 3D hand tracking, this paper also tests passive stereo for the state-of-the-art discriminative hand pose estimation\cite{sun2015cascaded}. The regression model is trained from the sequences in five of the backgrounds and tests in the remaining one.

The percentages of all joints that have a maximum error less than a threshold are computed for every sequence in \reffig{fig:individual_result} in which columns are for different backgrounds and rows are for different poses and tracking/estimation methods \cite{oikonomidis2011efficient,qian2014realtime,sun2015cascaded}. In order to easier compare different stereo matching methods and hand segmentations, the average percentages over different backgrounds are plotted in \reffig{fig:all_result}. The two rows in \reffig{fig:all_result} present the performance on simple counting and difficult random hand poses, respectively.

The red, green and blue curves in \reffig{fig:all_result} are obtained from Intel F200 \textbf{active depth camera}, the proposed stereo with the proposed hand segmentation method, and the traditional local stereo method with manual hand segmentation, respectively. As can be seen from every subfigure of \reffig{fig:all_result}, these three curves are very close to each other and the red curves (which correspond to the active depth camera) is the highest on average. \textit{This demonstrates that passive stereo is also suitable for 3D hand pose tracking and estimation as its performance is indeed comparable to the active depth cameras}. The blue curves in \reffig{fig:all_result} (c) can be even lower than the green curves because the hand disparities estimated from the proposed stereo matching method are normally better than the traditional stereo method although the manual hand segmentation is more robust than the proposed hand segmentation method.

The dark curves are obtained from the traditional local stereo method with the proposed hand segmentation method. According to \reffig{fig:all_result}, it is normally much lower than the proposed stereo method (green curves) since it cannot provide accurate hand mask which leads to less accurate hand pose tracking/estimation. \textit{This quantitatively demonstrates that the proposed stereo method has a much higher performance than the traditional stereo methods when used for 3D hand pose tracking/estimation}. Besides tracking accuracy, PSO converges slowly especially in random pose sequences which contain large global rotations. These are two major reasons that PSO fails to track hand poses in random pose sequences with traditional stereo. From \reffig{fig:individual_result}, ICPPSO cannot track hand poses in \textit{B3}, \textit{B4} and \textit{B5} for both counting and random poses with traditional stereo since skin color model is not as accurate as the rest sequences (see \reffig{fig:mask_comp} (c)) and ICP is not robust to inaccurate hand depth. CHPR cannot perform well with traditional stereo since the features extracted from inaccurate hand depth (getting from hand disparity) are also inaccurate.

The brown curves are obtained from the proposed stereo method with generic skin color model\cite{jones2002statistical} for hand detection in both \reffig{fig:all_result} (average percentage) and \reffig{fig:individual_result} (individual percentage). They exclude the video sequences captured with \textit{B4} and \textit{B6} backgrounds as the generic skin color model is completely fail on these video sequences and cannot detect the hand region. Brown curves are relatively lower than the green curves and it demonstrates the effectiveness of the proposed hand segmentation method. With generic skin color model, PSO has poor performance in random pose sequences. The reason is similar to PSO in traditional stereo matching with proposed hand segmentation. ICPPSO can work well for counting pose except for \textit{B3} with specularity regions in background (third row of \reffig{fig:mask_comp} (b)). The generic color model treats specularity as skin and stereo matching cannot estimate correct disparity in specularity regions. The disparities in specularity regions influence ICP which is the reason ICPPSO has poor performance for \textit{B3} counting pose. In the random pose, ICPPSO cannot work well for \textit{B1} since the shadow is too dark in hand when it faces downward and generic color model doesn't treat it as a hand. Also, ICP is not robust in this scenario. CHPR has comparable estimation performance with proposed stereo only in \textit{B5} since generic color model works well in \textit{B5} (fifth row of \reffig{fig:mask_comp} (b)) and the hand disparity is noisy in the rest backgrounds.

Meshstereo \cite{zhang2015meshstereo}, which is the state-of-the-art stereo algorithm on Middlebury benchmark, is also numerically compared with the proposed stereo for hand pose tracking/estimation in \reffig{fig:all_result} and \reffig{fig:individual_result}. Background \textit{B2} is ignored because Meshstereo gives very bad disparity estimates around the hand region with \textbf{author published source code and default parameters}. The adopted hand tracking/estimation algorithms do not work in this case. For the other backgrounds, the tracking/estimation performance (presented as the pink curves) is much worse than the proposed stereo. Apparently, Meshstereo is not suitable for 3D hand pose tracking/estimation.

The hand pose models estimated from different 3D hand pose tracking/estimation methods are visualized in \reffig{fig:est_result}(c)-(e). They visually agree with the segmented hand disparity maps in \reffig{fig:est_result}(b) under different backgrounds in \reffig{fig:est_result}(a).

From the experiments conducted in this section, we can conclude that when used for 3D hand tracking,
\begin{itemize}
  \item \textbf{the performance of passive stereo is low without correct hand segmentation;}
  \item \textbf{the performance of the proposed stereo matching method is close to active depth cameras.}
\end{itemize}

\section{Conclusions}
In this paper, a benchmark for evaluating hand pose tracking/estimation algorithms on passive stereo is developed. Unlike existing benchmarks, it contains both stereo images from a binocular stereo camera and depth images captured from an active depth camera. It has a total of 12 video sequences. Each sequence has 1,500 frames and thus can be used for long-term tracking evaluation. A novel stereo-based hand segmentation algorithm specially designed for hand tracking/estimation is proposed and the performance is demonstrated to be comparable to the active depth cameras under different challenging indoor scenarios. We also captured several sequences for outdoor hand pose tracking/estimation. It is difficult to obtain the ground truth labeling (because active depth sensor cannot work well under outdoor environment), and thus we do not show the quantitative performance of outdoor scenario in this paper. We will investigate into this labeling problem in the near future.

\begin{figure}[!ht]
\renewcommand{\tabcolsep}{0.5 pt}
\def\shfailure{0.34\linewidth}
\begin{center}
    \begin{tabular}{cc}
        \includegraphics[height=\shfailure]{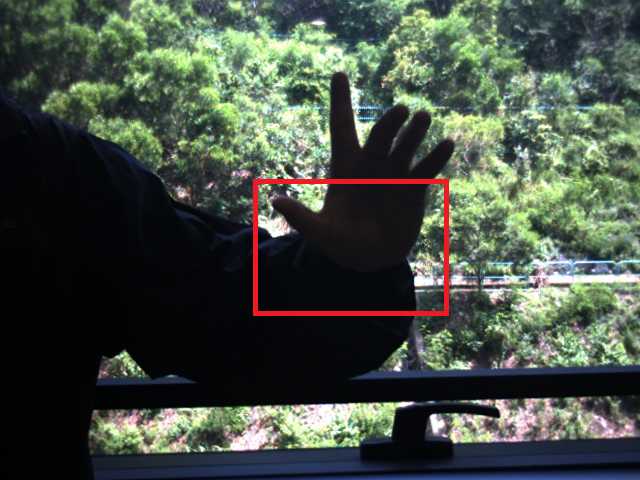} &
        \includegraphics[height=\shfailure]{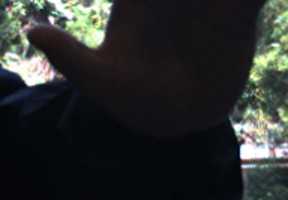}\\
        \small{(a) Color} & \small{(b) Close-up of (a)}\\
        \includegraphics[height=\shfailure]{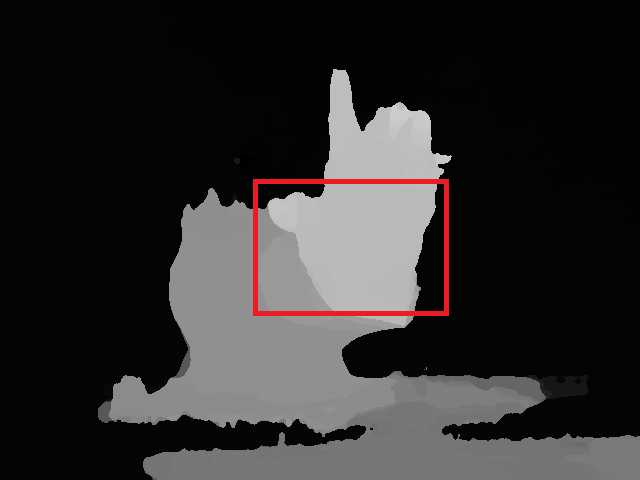} &
        \includegraphics[height=\shfailure]{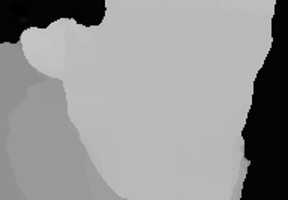}\\
        \small{(c) Proposed stereo} & \small{(d) Close-up of (c)}
    \end{tabular}
\end{center}
   \caption{A failure case due to bad lighting condition.}
\label{fig:failure}
    %\vspace{-6mm}
\end{figure}

The accuracy of passive stereo could drop dramatically under bad lighting environments. In this case, stereo matching noise is consequentially higher which results in lower-quality disparity estimates. Such a failure case is presented in \reffig{fig:failure}. No single global exposure can preserve both the colors of the outdoor scene and the details of the indoor object in \reffig{fig:failure}(a)-(b). The foreground indoor objects (including the hand and the arm) are too dark and thus lack of textures. Passive stereo is known to be vulnerable to this problem due to high-level matching noise. The disparity map estimated from the proposed method is shown in \reffig{fig:failure} (c)-(d). Note that the depth details/contrast in (d) are missing. We have not yet found a good solution except for installing LED lights on the camera.

%inside the the bottom region of hand spills out to that of the arm region. This leads to a wrong estimated disparity, which has a great influence on the following hand tracking/estimation.

\bibliographystyle{IEEEtran}
\bibliography{reference}

\end{document}